\documentclass[preprint,12pt]{elsarticle}
\usepackage{geometry}
\usepackage{graphicx}
\usepackage{amssymb}
\usepackage{lineno}
\usepackage{algorithm}
\usepackage{algorithmic}
\usepackage{float}
\usepackage{caption}%
\usepackage{amsmath}
\usepackage{siunitx}
\usepackage{pifont}% http://ctan.org/pkg/pifont
\newcommand{\xmark}{\ding{55}}%
\usepackage{amsfonts}
\usepackage{subfig}
\usepackage{booktabs}
\usepackage{placeins}
\usepackage{xr}
\usepackage{amsthm}
\usepackage{bbding}
\usepackage{array}
\usepackage{natbib}
\usepackage{relsize}
\usepackage{subfig}
\usepackage{graphicx}
\usepackage{esint}
\usepackage{tikz}
\usepackage{arydshln}
\usepackage{eurosym}
\usepackage{color}
\usepackage{multirow}
\usepackage{textcomp}
\usepackage{tabularx}
\usepackage{placeins}
\usepackage{lscape}
\usepackage{pdflscape}
\usepackage{amsmath}
\usepackage{leftidx}
\usepackage{float}
\usepackage{diagbox}
\usepackage{amssymb}
\usepackage{pgf,tikz,pgfplots}
\pgfplotsset{compat=1.14}
\usepackage{mathrsfs}
\usetikzlibrary{arrows}
\floatstyle{plaintop}
\restylefloat{table}
\usepackage{array}
\newcolumntype{L}[1]{>{\raggedright\let\newline\\\arraybackslash\hspace{0pt}}m{#1}}
\newcolumntype{C}[1]{>{\centering\let\newline\\\arraybackslash\hspace{0pt}}m{#1}}
\newcolumntype{R}[1]{>{\raggedleft\let\newline\\\arraybackslash\hspace{0pt}}m{#1}}

\usepackage[hidelinks=True,
            colorlinks = true,
            linkcolor = blue,
            urlcolor  = blue,
            citecolor = blue,
            anchorcolor = blue]{hyperref}
\setcitestyle{open={(},authoryear,close={)}}

\journal{Journal name}

\begin{document}
\definecolor{dtsfsf}{rgb}{0.8274509803921568,0.1843137254901961,0.1843137254901961}
\definecolor{rvwvcq}{rgb}{0.08235294117647059,0.396078431372549,0.7529411764705882}
\definecolor{dtsfsf}{rgb}{0.8274509803921568,0.1843137254901961,0.1843137254901961}
\definecolor{sexdts}{rgb}{0.1803921568627451,0.49019607843137253,0.19607843137254902}
\definecolor{rvwvcq}{rgb}{0.08235294117647059,0.396078431372549,0.7529411764705882}
\definecolor{wrwrwr}{rgb}{0.3803921568627451,0.3803921568627451,0.3803921568627451}
\begin{frontmatter}

%% Title, authors and addresses

%% use the tnoteref command within \title for footnotes;
%% use the tnotetext command for the associated footnote;
%% use the fnref command within \author or \address for footnotes;
%% use the fntext command for the associated footnote;
%% use the corref command within \author for corresponding author footnotes;
%% use the cortext command for the associated footnote;
%% use the ead command for the email address,
%% and the form \ead[url] for the home page:
%%
%% \title{Title\tnoteref{label1}}
%% \tnotetext[label1]{}
%% \author{Name\corref{cor1}\fnref{label2}}
%% \ead{email address}
%% \ead[url]{home page}
%% \fntext[label2]{}
%% \cortext[cor1]{}
%% \address{Address\fnref{label3}}
%% \fntext[label3]{}

\title{Trajectory Planning for Connected and Automated Vehicles: Cruising, Lane Changing, and Platooning}
\author{Xiangguo~Liu, Guangchen~Zhao,
        Neda~Masoud$^{*}$, Qi~Zhu}% <-this % stops a space

%\thanks{Xiangguo Liu and Qi Zhu are with Northwestern University, Evanston, IL 60208,  USA (e-mail: xg.liu@u.northwestern.edu,  qzhu@northwestern.edu). Guangchen Zhao is with the University of Maryland, College Park, MD 20742, USA  (email:gczhao@umd.edu). Neda Masoud is with the University of Michigan, Ann Arbor, MI 48109, USA  (email:nmasoud@umich.edu) }}
%% use optional labels to link authors explicitly to addresses:
%% \author[label1,label2]{<author name>}
%% \address[label1]{<address>}
%% \address[label2]{<address>}

%\address[a]{Department of Civil and Environmental Engineering, University of Michigan, AnnArbor, MI 48109, United States}

\begin{abstract}
Autonomy and connectivity are considered among the most promising technologies to improve safety, mobility, fuel and time consumption in transportation systems. Some of the fuel efficiency benefits of connected and automated vehicles (CAVs) can be realized through platooning. A platoon is a virtual train of CAVs that travel together following the platoon head, with small gaps between them. Vehicles may also reduce travel time by lane changing. In this paper, we devise an optimal control-based trajectory planning model that can provide safe and efficient trajectories for the subject vehicle and can incorporate platooning and lane changing. We embed this trajectory planning model in a simulation framework to quantify its efficiency benefits as it relates to fuel consumption and travel time, in a dynamic traffic stream. Furthermore, we perform extensive numerical experiments to investigate whether, and the circumstances under which, the vehicles in upstream of the subject vehicle may also experience second-hand fuel efficiency benefits.
\end{abstract}

\begin{keyword}
Autonomous Vehicles \sep Trajectory Planning \sep Connected Vehicles \sep Platoon Formation \sep Lane Changing \sep Mixed Traffic \sep Fuel Efficiency.

%% keywords here, in the form: keyword \sep keyword

%% PACS codes here, in the form: \PACS code \sep code

%% MSC codes here, in the form: \MSC code \sep code
%% or \MSC[2008] code \sep code (2000 is the default)

\end{keyword}
\cortext[cor1]{Corresponding author\\ E-mail address: nmasoud@umich.edu}
\end{frontmatter}

\section{Introduction}\label{sec:sec_1}
\noindent It is envisioned that in the near future transportation systems would be composed of vehicles with varying levels of connectivity and autonomy. Connected vehicle (CV) technology facilitates communication among vehicles, the infrastructure, and other road users \cite{doi:10.3141/2391-01}, allowing vehicles to see beyond the drivers' line of sight, and the transportation infrastructure to be proactive in responding to stochastic changes in road conditions and travel demand.

Autonomous vehicle technology enables automation of vehicles at different levels, where level 0 automation indicates no automation, and automation levels 1 and 2 refer to a single and multiple driving assistant systems being present in the vehicle, respectively. Level 3 automation allows the transfer of control authority between the human driver and the autonomous entity when the automation fails. Level 4 autonomy allows for the vehicle to control all functionalities within specified regions. Finally, in level 5 autonomy, vehicles can travel anywhere without any intervention from human drivers \cite{sae2014taxonomy}. 

Although each of the connected and automated vehicle technologies can be deployed independently in a vehicle, when combined they can provide a synergistic effect that goes beyond the sum of their individual benefits. It is expected that upon deployment, the connected and automated vehicle (CAV) technology could significantly improve mobility, enhance traffic flow stability, reduce congestion, and improve fuel economy, among other benefits. The degree to which such benefits can be realized in real-world conditions depends on a wide array of factors, among which trajectory planning of CAVs plays a major role \cite{Gasparetto2015Path}. The main purpose of trajectory planning is to provide a vehicle with a collision-free path, considering the vehicle dynamics, the surrounding traffic environment, and traffic rules \cite{6722462}. More advanced trajectory planning techniques could incorporate secondary objectives such as achieving fuel economy \cite{Zeng2018Globally,Han2014On,Yu2016Model,Lee2011Intelligent}.

Traditionally, trajectory planning has been mainly based on vehicle dynamics constraints, such as acceleration range, steering performance, etc. More advanced driving assistance systems (ADAS), e.g., adaptive cruise control (ACC), enhance trajectory planning through utilizing data collected by vehicle on-board sensors.  CV technology provides an opportunity to incorporate more diverse types of data (e.g., weather conditions) from a wider spatial range (e.g., from vehicles beyond the line of sight of the vehicle's on-board sensors). However, there is a need to develop algorithmic tools that can incorporate this information into trajectory planning. Several attempts, such as Connected Cruise Control (CCC) \cite{Zhang2013Designing, Orosz2016Connected} and Cooperative Adaptive Cruise Control (CACC) \cite{Zhang2018Hierarchical, Ge2017Optimal, 7314936, MILANES2014285} have been made to incorporate vehicle-to-vehicle (V2V) communications into trajectory planning. CACC is one of the most promising technologies that allow CVs to autonomously, and without need for a central management system, plan their trajectories using V2V communications \cite{wischhof2005information}. The information flow topology in a CACC system typically includes predecessor following, predecessor-leader following, bidirectional topology, etc. \cite{zheng2014influence}. Advanced communication protocols, such as Dedicated Short Range Communications (DSRC), LTE, and 5G are proposed and developed to improve the communication bandwidth of V2V communications \cite{Fan2006Reliability, Uhlemann2016Connected, S2017Millimeter}. 

\begin{landscape}
\begin{table}[h]
\footnotesize
\caption{Overview of the trajectory planning literature}
\label{tab:lit}
%\scalebox{0.75}{
\begin{tabular}{l L{4.5cm} c c L{2.5cm} ccc C{1.5cm} }
\hline
\multirow{2}{*}{\textbf{Study}} & \multirow{2}{*}{\textbf{Environmental dynamics}}                  & \multirow{2}{*}{\textbf{Platoon}} & \multirow{2}{*}{\textbf{Lane changing}} & \multirow{2}{*}{\textbf{Mixed traffic}} & \multicolumn{4}{c}{\textbf{Cost}}                  \\
                                       &       &                                                     &                                   &                                         & tracking  & fuel       & time      & comfort
                                       /safety                                  \\
 \hline                                   
\cite{Gu2016On}                      & pedestrian/static object                           & \xmark   & \checkmark                                 & pedestrian/static object                & \checkmark          & \xmark          & \xmark         & \checkmark                                      \\
\cite{Zeng2018Globally}                      & stop signs/ traffic lights                         & \xmark                                 & \xmark               & \xmark                        & \xmark          & \checkmark          & \xmark         & \xmark                                      \\
\cite{Fu2015A}                       & \begin{tabular}[c]{@{}l@{}}multiple dynamic obstacles \\ (3 surrounding vehicles) \end{tabular}  & \xmark         & \checkmark                         & \xmark                                       & \xmark          & \checkmark          & \checkmark         & \checkmark                                 \\
\cite{8410769}                     & \begin{tabular}[c]{@{}l@{}}fixed velocity of \\ surrounding vehicles\end{tabular}             & \checkmark                & \checkmark                   & connected/non-connected                 & \checkmark          & \xmark          & \xmark         & \xmark                               \\
\cite {8317585}                        & \begin{tabular}[c]{@{}l@{}}curvy road segment with one \\ obstacle\end{tabular}               & \xmark           & \xmark                      & \xmark                                       & \xmark          & \xmark          & \xmark         & \checkmark                                      \\
\cite{NTOUSAKIS2016464}               & \begin{tabular}[c]{@{}l@{}}fixed velocity of the leading \\ vehicle\end{tabular}              & \checkmark           & \checkmark                         & all automated vehicles                  & \xmark          & \checkmark          & \checkmark         & \checkmark                                  \\
\cite{Gritschneder2018Fast}             & \xmark         & \xmark                                           & \xmark                                 & \xmark                                       & \checkmark          & \xmark          & \xmark         & \xmark                                       \\
\textbf{This paper}                      & \textbf{all vehicles}                       & \textbf{\checkmark}        &  \checkmark             & \textbf{human driver/CAV}              & \xmark & \textbf{\checkmark} & \textbf{\checkmark}& \textbf{\checkmark}      \\
\hline
\end{tabular}
\end{table}
\end{landscape}

Table \ref{tab:lit} summarizes the recent studies in the literature that have focused on trajectory planning of CAVs, with different levels of automation. This table points out multiple attributes of these studies, including the degree to which they are reactive to the dynamics of the traffic stream, their capability to model vehicles with various levels of autonomy and connectivity, whether they are capable of platoon formation, and their cost functions. The rest of this section elaborates on the specifics of these attributes.

The ultimate goal of trajectory planning is to enable vehicles to travel safely and efficiently in real traffic conditions. Therefore, different trajectory planning algorithms are developed for implementation in different contexts, to capture different abstractions of real-world conditions, e.g., obstacles, curved roads, signal lights, and mixed traffic components, etc. In \cite{Gu2016On}, Gu et al. focus on the subject vehicle's movement around a single static obstacle, and its distance-keeping and overtaking of a single leading vehicle. \cite{Zeng2018Globally} propose a dynamic programming (DP) algorithm for speed planning in a transportation network with stop signs and traffic lights. \cite{8317585} presents a method that exploits the complete permissible road width in curvy road segments to increase driving comfort and safety through minimized steering actuation. \cite{Fu2015A} develops a hybrid intelligent optimization algorithm based on ordinal optimization for autonomous vehicles traveling in environments with multiple moving obstacles. \cite{8410769} and  \cite{NTOUSAKIS2016464} consider the impact of surrounding vehicles with fixed velocity on trajectory planning of the subject vehicle. In general, the degree to which different models are set to imitate real traffic conditions depends on research priorities. The closer the environment is to real-world conditions, the higher the accuracy and reliability of trajectory planning, but the higher the computational complexity and the worse the real-time performance. As such, literature is in general very limited in capturing the dynamics of the driving environment.

%The proximity of the environmental conditions to reality depends on the different research priorities. 

%Their proposed DP algorithm is not a real-time method, as it requires considerable computational resources. However, its global optimal results can be used to evaluate the performances of other speed planning methods.

%lane-changing
Lane changing is another important component of trajectory planning. Lane changing is one of the most challenging driving maneuvers for researchers to understand and predict, and one of the main causes of congestion and collisions in the transportation system \cite{TALEBPOUR2015420}. The real-time information received from the driving environment and other road users can be used to facilitate lane changing maneuvers that enhance safety, comfort and traffic efficiency \cite{Luo2016A}. %Researches about the lane changing process have been conducted from different aspects. 
The developments in lane changing models before 2014 are comprehensively reviewed in \cite{rahman2013review, Zheng2014Recent}. Zheng et al. \cite{Zheng2014Recent} claim that in real lane changing situations, drivers can simultaneously monitor and evaluate multiple spacings in the target lane and make a decision on where and how to execute the lane change. In \cite{TALEBPOUR2015420}, two types of games, under complete information in the presence of connected vehicle technology, and under incomplete information in its absence, are proposed for modeling lane changing behavior. Simulation results indicate that the game theoretic-based lane changing models are more realistic than the basic gap-acceptance model and the MOBIL model. Wang et al. \cite{WANG201573} propose a predictive model for lane changing control that considers both discrete lane changing decisions and continuous acceleration values. The lane changing method proposed by Luo et al. in 2016 \cite{Luo2016A} executes lane changing maneuvers; however, their model is not capable of making lane-changing decisions. Through the above review of the research on lane changing behavior, it is particularly noteworthy that most research on the lane changing maneuver only focuses either on when or where to change lane, or on the execution of lane changing after the decision of changing lane has been made. In real life, both these decisions, i.e., decision-making on where and when to change lane and the lane-changing execution, are made simultaneously.

%Platooning
Platooning makes one of the most interesting and important components of trajectory planning in the next generation of transportation systems. The capability to incorporate platooning is another factor that differentiates existing trajectory planning methods.
Platooning is a specific application of the CV technology that can introduce a wide range of vehicle- and system-level benefits. A platoon is a single-file line (i.e., a virtual train) of vehicles that, owing to constant communication, are able to travel with small gaps between them. Platoon formation can introduce many benefits including (i) energy efficiency through reducing the aerodynamic drag force on platoon members \cite{Zabat1995The,Alam2015Heavy}, as well as reducing emissions \cite{Farnsworth2001EL}; (ii) increasing road capacity through reducing the headways between vehicles; (iii) reducing stochasticity in the traffic stream by having platoon members follow the platoon head, thereby reducing the likelihood of highway traffic breakdown, improving travel times, and increasing travel time reliability \cite{Shladover2015COOPERATIVE, LIORIS2017292}; and (iv) facilitating real-time management of traffic and improving mobility by aggregating the unit of traffic from an individual vehicle to a cluster of vehicles. TABLE \ref{tab:lit} specifies the studies in the literature that incorporate platooning. Note that a check mark for the field `platoon' in this table indicates the capability of platoon formation, rather than platoon control strategies \cite{Contet2009Bending,Guo2011Hierarchical} or intra-platoon communication \cite{El2012Vehicle,Maiti2017A}.

The ability to capture the heterogeneity in the level of connectivity and autonomy of vehicles is another factor that differentiates existing trajectory planning methods, as described in Table \ref{tab:lit}.
Finally, trajectory planning methods are different in terms of their objective function. In general, the goal is to find the least-cost trajectory, where the cost function could include any combination of the following factors: time-cost of the trip (i.e., trip length), fuel consumption, comfort and safety of on-board passengers, and precision in tracking (i.e., the degree to which the vehicle deviates from a pre-specified ideal trajectory). 

To the best of our knowledge, there is not yet a study that comprehensively discusses trajectory planning and lane changing for a connected and automated vehicle in a fully dynamic environment with mixed traffic, considering platoon formation for both the subject vehicle and its surrounding vehicles. The goal of this paper is to determine a lane changing and platoon-joining policy that can minimize the generalized cost over a prediction horizon for the subject vehicle, which is considered to be a connected and automated vehicle, in a dynamic environment with mixed traffic. Here a `dynamic' environment refers to an environment in which the vehicles surrounding the subject vehicle may enter or exit the traffic stream, merge into or split from a platoon, change lane, and adjust their velocities. %Along the trip, the surrounding environment of the subject vehicle is dynamically changing. 
The joint lane changing and platoon-joining decisions are made by the subject vehicle in such a dynamic traffic environment so as to minimize a generalized cost function. We use simulations to assess whether and how, in general, optimal control of a single vehicle can affect the fuel consumption of its surrounding vehicles. These experiments allow us to quantify whether, and the extent to which, having a few CAVs in the traffic stream could have fuel efficiency benefits that go beyond the CAVs themselves.

The rest of the paper is organized as follows: First, we formulate an optimal control model for planning the trajectory of a CAV. Next, we present a simulation environment that consists of a two-lane highway with multiple on- and off-ramps and a dynamic traffic stream. In particular, we describe how vehicles with various levels of autonomy interact with each other in the simulation environment. Finally, in section \ref{sec:NUMERICAL EXPERIMENTS}, we conduct a series of analyses under various traffic conditions to quantify the fuel-efficiency benefits of our approach for the subject vehicle as well as those of its surrounding vehicles within platoons and as free agents. We end the paper by summarizing the takeaways.
    %we present a literature review respectively of related works; Second,we describe the design of the trajectory planning and platoon joining based on the optimal control theory; Then we provide the results of simulations and experiments on such a model in several typical scenarios; Finally, we present the conclusions with some summary remarks and future research needs.

%{\color{blue}Xiangguo: It's a mixed traffic environment, all kinds of vehicles are considered (connected vehicle or non-connected vehicle, automated vehicle or human-driving vehicle). Our goal is to make decisions for the subject vehicle (a connected and automated vehicle) regarding lane changing and platooning along a trip from origin to destination, in a dynamic environment. Here a dynamic environment means surrounding vehicles can enter or exit from the highway by on-ramps and off-ramps, merge into for split from a platoon, change lane and adjust velocity. Along the trip, the surrounding environment of the subject vehicle is dynamically changing.

%adjust--adjust the final positions of all actions to be the same, the cost would be adjusted accordingly, so that the overall costs can be compared. find--find the least cost and the associated action

%}

\section{METHODS}
The goal of this study is to design an optimal control-based trajectory planning model that can be utilized by an automated (level 2 or higher autonomy) vehicle, hereafter referred to as the ``subject vehicle''. The optimal control model will be designed to incorporate microscopic traffic information from the traffic stream in the local neighborhood of the subject vehicle, with the goal of devising fuel and time efficient trajectories that may include merging into a platoon and changing lane. We start this section by describing the optimal control model. We then describe a simulation environment that we will use to quantify the overall cost savings for the subject vehicle as well as its surrounding traffic.

\subsection{Optimal Control Model}
In this section, we devise an optimal control model to determine the trajectory of the subject vehicle in real-time. The proposed optimal control model is provably safe, and is designed to account for fuel and time efficiency as well as comfort of on-board passengers.

The control parameters of the optimal control model include lateral and longitudinal acceleration/deceleration and platoon membership. While adjusting acceleration can be considered as a single action that can be almost instantaneously carried out, a change in lane position and platoon membership is a lengthier process and may require multiple sub-actions, as described in Table \ref{tab:seq}. As demonstrated in this table, at each point in time the subject vehicle can be in one of the following six states: ($i$) `left lane; free agent', indicating that the vehicle is in left lane and is not part of any platoon, ($ii$) `right lane; free agent', indicating that the vehicle is in right lane and is not part of any platoon, ($iii$) `left lane; in platoon (active)', indicating that the subject vehicle is in the left lane and is the platoon lead, and the scheduled platoon splitting position has not yet reached, ($iv$) `right lane; in platoon (active)', indicating that the subject vehicle is in right lane and is the platoon lead, and the scheduled platoon splitting position has not yet reached, ($v$) `left lane; in platoon (passive)', indicating that the subject vehicle is in left lane, the platoon splitting position has reached, and the platoon the subject vehicle was formerly leading is in the process of dissolving, and ($vi$) `right lane; in platoon (passive)', indicating that the subject vehicle is in right lane, the platoon splitting position has reached, and the platoon the subject vehicle was formerly leading is in the process of dissolving. 

Table \ref{tab:seq} shows that at each point in time, the subject vehicle switch from its current state to a target state. Depending on its initial and target states, the subject vehicle may need to complete a sequence of sub-actions, including `wait', `merge', `split' and `lane change'. The `wait' sub-action indicates that the vehicle needs to maintain its state after completing its previous sub-action. The sub-actions `merge' and `split' indicate merging into a platoon and splitting from a platoon, respectively. Finally, the `lane change' sub-action indicates changing to the other lane. For example, if the target state `right lane; in platoon' is the selected action under the current state `left lane; in platoon (active)', then the subject vehicle needs to complete the sequence of sub-actions `split$\longrightarrow$wait$\longrightarrow$lane change$\longrightarrow$merge$\longrightarrow$wait'.

\subsubsection{The trajectory function}
Following \cite{Luo2016A}, we use a quintic function, based on time, as our trajectory function for each sub-action. The quintic function is selected because it guarantees a smooth overall trajectory, even with multiple different sub-actions. Eq. (\ref{eq:traj}) shows the trajectory function,
\begin{equation} \label{eq:traj}
%x(t)=\sum^{N_{\text{act}}}_{i=1}(a^{i}_{5}t^{5}+a^{i}_{4}t^{4}+a^{i}_{3}t^{3}+a^{i}_{2}t^{2}+a^{i}_{1}t+a^{i}_{0})f_i(t)
\begin{cases}
x(t)=\sum^{N_{\text{act}}}_{i=1}(a^{i}_{5}t^{5}+a^{i}_{4}t^{4}+a^{i}_{3}t^{3}+a^{i}_{2}t^{2}+a^{i}_{1}t+a^{i}_{0})f_i(t)
\\
y(t)=\sum^{N_{\text{act}}}_{i=1}(b^{i}_{5}t^{5}+b^{i}_{4}t^{4}+b^{i}_{3}t^{3}+b^{i}_{2}t^{2}+b^{i}_{1}t+b^{i}_{0})f_i(t)
\end{cases}
\end{equation}
where $N_{\text{act}}$ is the number of sub-actions the subject vehicle needs to complete. Function $f_i(t)$ may be formulated as
\begin{equation}
f_i(t)=
\begin{cases}
\begin{aligned}
1                      &&& t_{i-1} \leq t < t_i
\\
0                      &&& \text{otherwise}
\end{aligned}
\end{cases}
\end{equation}
where $[t_{i-1} ~,~ t_i]$ is the time window for completing the $i$'th sub-action, and $t_{N_{\text{act}}}$ is the prediction horizon.

\subsubsection{Boundary conditions}
For every sub-action, the following boundary conditions must be satisfied,
\begin{equation}\label{starting point}
\begin{cases}
x(t_{i-1})=x_{t_{i-1}} \text{, ~} \dot{x}(t_{i-1})=v_{x,t_{i-1}} \text{,~ } \ddot{x}(t_{i-1})=a_{x,t_{i-1}},
\\
y(t_{i-1})=y_{t_{i-1}} \text{, ~} \dot{y}(t_{i-1})=v_{y,t_{i-1}} \text{,~ } \ddot{y}(t_{i-1})=a_{y,t_{i-1}}
\end{cases}
%\begin{cases}
%x(t_s)=x_{t_s} \text{, } \dot{x}(t_s)=v_{x,t_s} \text{, } \ddot{x}(t_s)=a_{x,t_s}
%\\
%y(t_s)=y_{t_s} \text{, } \dot{y}(t_s)=v_{y,t_s} \text{, } \ddot{y}(t_s)=a_{y,t_s}
%\end{cases}
\end{equation}

\begin{equation}\label{ending point}
\begin{cases}
x(t_i)=x_{t_i} \text{, ~} \dot{x}(t_i)=v_{x,t_i} \text{, ~} \ddot{x}(t_i)=a_{x,t_i}
\\
y(t_i)=y_{t_i} \text{, ~} \dot{y}(t_i)=v_{y,t_i} \text{, ~} \ddot{y}(t_i)=a_{y,t_i}
\end{cases}
%\begin{cases}
%x(t_e)=x_{t_e} \text{, } \dot{x}(t_e)=v_{x,t_e} \text{, } \ddot{x}(t_e)=a_{x,t_e}
%\\
%y(t_e)=y_{t_e} \text{, } \dot{y}(t_e)=v_{y,t_e} \text{, } \ddot{y}(t_e)=a_{y,t_e}
%\end{cases}
\end{equation}
where $t_{i-1}$ and $t_i$ are the starting and ending time for the $i$th sub-action, respectively, %As shown in \eqref{starting point}, 
and $x_{t_{i-1}}$, $v_{x,t_{i-1}}$, $a_{x,t_{i-1}}$, $y_{t_{i-1}}$, $v_{y,t_{i-1}}$ and $a_{y,t_{i-1}}$ are the longitudinal and lateral geo-coordinates, velocity, and acceleration for the starting point of the sub-action, respectively. These values are accordant with the ending point for the last sub-action. For each sub-action, the longitudinal coordinate, velocity and acceleration at the end of the sub-action as well as the duration of the sub-action are all free variables that are optimized.

\begin{table*}[h]
\centering
\caption{Sub-action sequences for each state-action tuple}
%{\color{blue}[Third state, first action: The vehicle stays in the left lane, but wants to split from the platoon. First, why would this happen at all? Second, even if it is a plausible action, if the vehicle splits and waits (as shown in the arrangement section) it is still part of a platoon (it is the lead vehicle in a new platoon)] [What is active and passive in the state column? there is no difference in the arrangements for them.]}}

% I just list all the possibilities, the cost computed will decide which one should the subject vehicle choose. For example, the subject vehicle wants to be a free agent near the exit. 

% It is assumed that once the subject vehicle split, the vehicle behind the subject vehicle would also split, thus the subject vehicle becomes a free agent. Since we only control the motion of the subject vehicle, we don't analyze the splitting behavior of the following vehicle. This assumption is helping to make the two situation similar: the subject vehicle is in the middle of platoon; the subject vehicle is the last one.

% In one word, passive means $d=0$, the other vehicles would exit in this road piece, so they have to split from the subject vehicle and the subject vehicle has no other choice. Active means that the subject vehicle can choose stay in the platoon or leave.
\label{tab:seq}
\scalebox{0.85}{
\begin{tabular}{l|l|l}
\hline
Initial state                                             & Target state                 & Sub-action sequence                                                                                                             \\ \hline
\multirow{4}{*}{left lane; free agent}            & left lane; free agent  & wait                                                                                                                     \\ %\cline{2-3} 
                                                  & left lane; in platoon  & merge$\longrightarrow$wait                                                                          \\ %\cline{2-3} 
                                                  & right lane; free agent & wait$\longrightarrow$lane change$\longrightarrow$wait                                                                  \\ %\cline{2-3} 
                                                  & right lane; in platoon & wait$\longrightarrow$lane change$\longrightarrow$merge$\longrightarrow$wait                       \\ \hline
\multirow{4}{*}{right lane; free agent}           & left lane; free agent  & wait$\longrightarrow$lane change$\longrightarrow$wait                                                                  \\ %\cline{2-3} 
                                                  & left lane; in platoon  & wait$\longrightarrow$lane change$\longrightarrow$merge$\longrightarrow$wait                       \\ %\cline{2-3} 
                                                  & right lane; free agent & wait                                                                                                                     \\ %\cline{2-3} 
                                                  & right lane; in platoon & merge$\longrightarrow$wait                                                                          \\ \hline
\multirow{4}{*}{left lane; in platoon (active)}   & left lane; free agent  & split$\longrightarrow$wait                                                                                               \\ %\cline{2-3} 
                                                  & left lane; in platoon  & wait                                                                                                                     \\ %\cline{2-3} 
                                                  & right lane; free agent & split$\longrightarrow$wait$\longrightarrow$lane change$\longrightarrow$wait                                            \\ %\cline{2-3} 
                                                  & right lane; in platoon & split$\longrightarrow$wait$\longrightarrow$lane change$\longrightarrow$merge$\longrightarrow$wait \\ \hline
\multirow{4}{*}{right lane; in platoon (active)}  & left lane; free agent  & split$\longrightarrow$wait$\longrightarrow$lane change$\longrightarrow$wait                                            \\ %\cline{2-3} 
                                                  & left lane; in platoon  & split$\longrightarrow$wait$\longrightarrow$lane change$\longrightarrow$merge$\longrightarrow$wait \\ %\cline{2-3} 
                                                  & right lane; free agent & split$\longrightarrow$wait                                                                                               \\ %\cline{2-3} 
                                                  & right lane; in platoon & wait                                                                                                                     \\ \hline
\multirow{4}{*}{left lane; in platoon (passive)}  & left lane; free agent  & split$\longrightarrow$wait                                                                                               \\ %\cline{2-3} 
                                                  & left lane; in platoon  & split$\longrightarrow$wait$\longrightarrow$merge$\longrightarrow$wait                                                    \\ %\cline{2-3} 
                                                  & right lane; free agent & split$\longrightarrow$wait$\longrightarrow$lane change$\longrightarrow$wait                                            \\ %\cline{2-3} 
                                                  & right lane; in platoon & split$\longrightarrow$wait$\longrightarrow$lane change$\longrightarrow$merge$\longrightarrow$wait \\ \hline
\multirow{4}{*}{right lane; in platoon (passive)} & left lane; free agent  & split$\longrightarrow$wait$\longrightarrow$lane change$\longrightarrow$wait                                            \\ %\cline{2-3} 
                                                  & left lane; in platoon  & split$\longrightarrow$wait$\longrightarrow$lane change$\longrightarrow$merge$\longrightarrow$wait \\ %\cline{2-3} 
                                                  & right lane; free agent & split$\longrightarrow$wait                                                                                               \\ %\cline{2-3} 
                                                  & right lane; in platoon & split$\longrightarrow$wait$\longrightarrow$merge$\longrightarrow$wait                                                    \\ \hline
\end{tabular}}
\end{table*}

\subsubsection{Constraint sets}
There are a number of constraints on the position, speed, acceleration, and jerk of the subject vehicle, elaborated in the following:

%(1) Position limitation: During the lane changing process, the subject vehicle should stay between the two lane centers, as stated in eq (\ref{eq:position1}), 

%\begin{equation}\label{eq:position1}
%y_{t_s}<y(t)<y_{t_s} + (-1)^\ell \: w 
%\end{equation}

%\noindent where $y(t)$ is the lateral position of the subject vehicle at time $t$ in the lane changing process, $w$ is the lane width, and $\ell$ is a binary indicator taking the value 1 if the subject vehicle is changing from left lane to right lane and the value 2 when the vehicle is changing from the right lane to the left lane. 
\vspace{0.5pc}
\noindent 1. \underline{Speed limitation}: The longitudinal speed of the subject vehicle should be no more than the maximum speed in its lane, and should always be non-negative, as presented in Eq. (\ref{eq:speed1}):

\begin{equation}\label{eq:speed1}
0\leq v_{x}(t)\leq v_{x,\max}^{l}
\end{equation}
where $v_{x}(t)$ denotes the longitudinal speed of the subject vehicle, $l$ indicates the lane in which the vehicle is traveling, and $v_{x,\max}^{l}$ denotes the maximum vehicle speed in lane $l$.

\vspace{0.5pc}
\noindent 2. \underline{Collision avoidance}: The subject vehicle should maintain a minimum time gap (denoted by $t_{\text{p}}$) from its immediate downstream vehicle during all sub-actions, as in Eq. (\ref{eq:collision1}),

\begin{equation}\label{eq:collision1}
\min \big( x_{L}(t)-x_{\text{sub}}(t) \big)  > t_{p}~  v_{\text{sub}}(t)
\end{equation}
where $x_{L}(t)$ is the position of the immediate downstream vehicle (i.e., the leader), and $x_{\text{sub}}(t)$ and $v_{\text{sub}}(t)$ are the position and velocity of the subject vehicle, respectively.

%\begin{equation}\label{eq:collision2}
%\min \big( x_{Ld}(t)-x_{M}(t) \big) >\tau  v_{Ld}(t)
%\end{equation}

%For lane changing sub-action, the distance between the subject vehicle and the following vehicle in destination lane should also follow the constraint (\ref{eq:collision3}).

%\begin{equation}\label{eq:collision3}
%\min\big( x_{M}(t)-x_{Fd}(t) \big) > T_p  v_{M}(t)
%\end{equation}

\vspace{0.5pc}
\noindent 3. \underline{Acceleration bound}: During all sub-actions, the longitudinal or lateral acceleration of the subject vehicle cannot exceed a maximum acceleration due to the mechanical performance limitations and safety considerations. This constraint is enforced in Eq. (\ref{eq:acceleration1}),

\begin{equation}\label{eq:acceleration1}
|a_{x,y}|<a_{\max}
\end{equation}

\noindent where $a_{\max}$ is the maximum acceleration.
% Ciuffo, Biagio, et al. "Capability of Current Car-Following Models to Reproduce Vehicle Free-Flow Acceleration Dynamics." IEEE Transactions on Intelligent Transportation Systems 19.11 (2018): 3594-3603.
\vspace{0.5pc}
\noindent 4. \underline{Jerk bound}: Since the subject vehicle's jerk directly influences the comfort level as well as the safety of on-board passengers, we bound the jerk by a maximum value as stated in Eq. (\ref{eq:jerk1}),

\begin{equation}\label{eq:jerk1}
|j_{x,y}|<j_{\max}
\end{equation}
\noindent where $j_{\max}$ is the maximum jerk.

%The optimal control model poses strict boundary conditions for each sub-action. As such, it may take an unrealistically long period for the subject vehicle to meet these conditions. Examples of such strict conditions include the requirement for the subject vehicle to obtain the exact same speed as the leading vehicle in the (wait) sub-action, or travel in the exact central lateral position in a lane. However, these conditions are not strictly required in real driving situations. As such, we set a parameter $tol$ to represent the tolerance for satisfying these conditions; that is, a sub-action would be considered as completed as long as its error does not exceed the tolerance.

\subsubsection{Objective function}
We consider fuel cost and time cost as the objective function in our optimization problem, as stated in Eq. (\ref{eq:cost1}) below,

\begin{equation}\label{eq:cost1}
\begin{aligned}
C_{\text{overall}}= & \eta_f ~\Sigma _{i=1}^{N_{\text{act}}} ~ \beta(i)~ \int _{t_{i-1}}^{t_i} \big(~ \gamma_{\text{AR}} v^{2}(t)+\gamma_{\text{RR}}+\gamma_{\text{GR}}+  \gamma_{\text{IR}}(a(t))_{+} \big)~v(t) ~ dt\\ & \qquad\qquad+\eta_t ~(t_{i}-t_{i-1})
\end{aligned}
\end{equation}

%\noindent where $t_{i-1}$ and $t_i$ denote the starting and ending times of the subject vehicle's trip through the $i$th sub-action, respectively. 

The four terms $\gamma_{\text{AR}} v^{2}(t)$, $\gamma_{\text{RR}}$, $\gamma_{\text{GR}}$, and $\gamma_{\text{IR}}(a(t))_{+}$ are the aerodynamic resistance force, rolling resistance force, grade resistance force and inertia resistance force, respectively. For detailed expressions of these forces, refer to \cite{gillespie1992fundamentals}. The parameter $\eta_f$ is the fuel cost for unit energy consumed by the vehicle, and is measured in dollars. The parameter $\eta_t$ is the time cost for a unit time consumed in this trip, and is measured in dollars. The parameter $\beta(i)$ indicates the fuel saving coefficient for sub-action $i$. We set $\beta(i)=1$ for a free agent, and $\beta(i)=0.9$ for a platoon member. Furthermore, we set $\beta(i)=0.95$ for split and merge sub-actions. The reason for assuming a $\beta$ value less than one for these sub-actions will be explained in the next section. 

\subsection{The Simulation Environment}
	
\begin{figure}[h]
    \centering\includegraphics[scale=0.42]{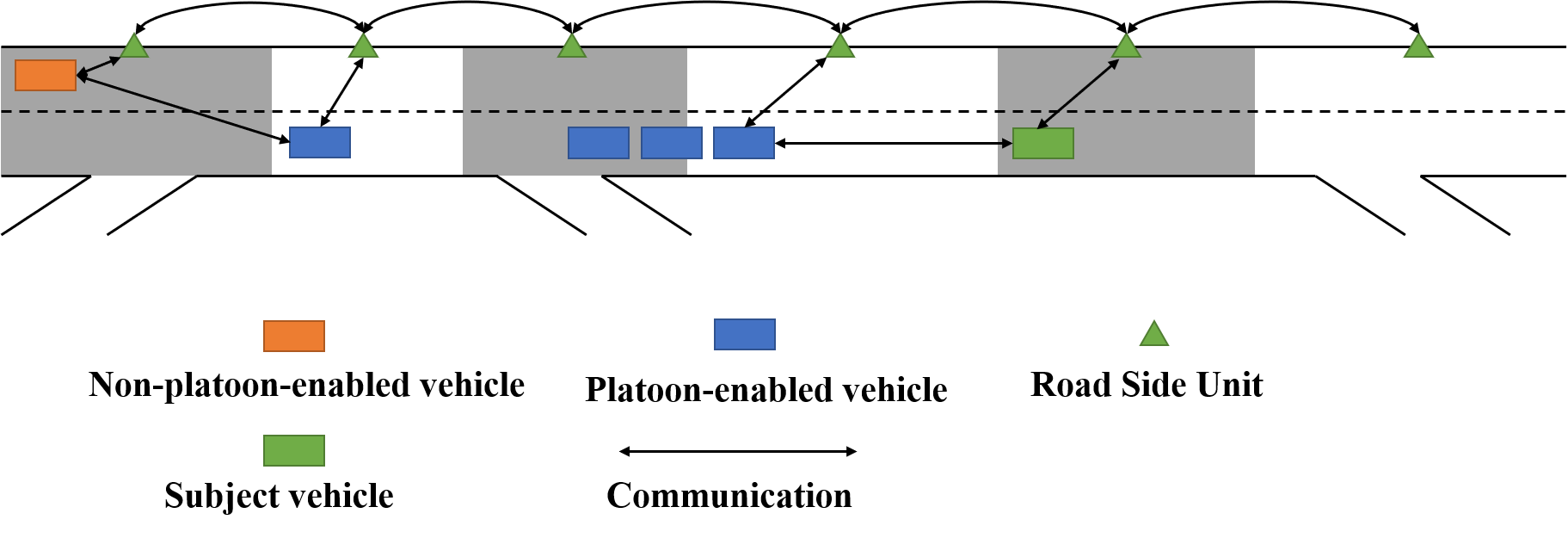}
    \caption{The communication and control framework}\label{fig:com&sen}
    \end{figure}

In this study, we consider a mixed traffic stream with various levels of autonomy. Specifically, we model both vehicles that are human-driven and not platoon-enabled, and platoon-enabled vehicles. A platoon-enabled vehicle is a vehicle that has level 2 or higher autonomy (and is equipped with distance sensing and keeping technology such as adaptive cruise control) according to the Society of Automotive Engineer's (SAE's) classification. 
Furthermore, in this study we assume that all vehicles are connected; that is, all vehicles can communicate with each other and with road side units (RSUs) using dedicated short range communication (DSRC) devices, with a reliable communication range of 300 meters. Figure \ref{fig:com&sen} demonstrates the communication and control framework of our work. 
	
	%We assume connectivity between any two vehicles is enabled by RSU and communication modules inside vehicles. We also assume there exists human-driving vehicles, platooned-enabled vehicles, automated vehicles at the same time. According to \trbcite{force2015white}, the terms in our work are defined as, human-driving vehicles are level 0 and 1, platooned-enabled vehicles are level 2 and 3, automated vehicles are level 4 and 5. Of course, automated vehicles are also platooned-enabled, but we just name it to show higher autonomy. We will also use regular vehicles in this paper, which are the same as human-driving vehicles. Thus the mixed traffic in our work is in terms of autonomy, rather than connectivity. 
	
	To develop a simulation environment for the system, we divide the transportation network into a number of ``road pieces''. A road piece is a section of a road that satisfies the following two conditions: ($i$) the macroscopic traffic conditions, to which we refer as ``traffic states'', are likely to be homogeneous within a road piece. For example, on a highway segment the traffic conditions around on- and off-ramps are typically different from their upstream and downstream segments, indicating that on- and off-ramps require dedicated road pieces; and %Note that although two road sections with different traffic states form two road sections, the reverse is not necessary through--two consecutive road sections could have the same traffic state.
	($ii$) vehicles within a road piece are able to communicate with each other, either directly or through RSUs. This requirement implies that in case of DSRC-enabled communication, the length of a road piece cannot exceed 600m so as to enable all vehicles to become connected using a single RSU located in the middle of the road piece. Limiting the length of a road piece ensures that, with strategic positioning of RSUs, all connected vehicles can receive microscopic traffic information of their neighbors (i.e., trajectories including longitude and latitude, velocity, acceleration, braking, steering angle, etc.), and use this information to plan more informed and efficient trajectories.
	
	%A road piece is considered to have similar traffic situation, which may be separated by on-ramp or off-ramp. Hence communication enabled by RSU, 5G or DSRC can provide the microscopic information of the traffic environment to the subject vehicle, which is an automated vehicle. The subject vehicle can thus estimate the motion of surrounding vehicles within a reliable communication range (300 meters for DSRC) in the future and make decisions accordingly. 

%	{\color{blue}We consider time delay in our work, which you cannot see in many other works(indicate specific papers)} 
	
	In our modeling of a traffic stream characterized with full connectivity and a heterogeneous level of autonomy, we account for the delay between the occurrence of a stimulus and the execution of an action in response to it.
	In case of a human being the driving authority, this delay is referred to as the perception-reaction time \cite{basak2013modeling}, and accounts for the perception delay (either by the driver or from the part of the vehicle sensors), the decision-making delay, and the execution delay. In case of the autonomous entity being in charge, this delay can be attributed to sensory delay, delay in the communication network, computational time, and actuation delay.
	
	%the period of time between the moment an stimulus requires a reaction, and the moment the reaction takes place--during this period the human driver perceives the stimulus, makes a decision on how to react, and persecutes the action. In cases where the autonomous vehicle is in charge, the delay 

	%Time delay exists in all three kinds of vehicles discussed in our work. For regular vehicles, time delay is mainly the reactive delay resulted by humans, while for automated vehicles, time delay is mainly the computation delay resulted by the computation process. Platooned-enabled vehicles may suffer either computation delay or reaction delay at different situations. Other delays, such as the communication delay or actuation delay is a small portion compared with the reaction delay or computation delay, which are not considered in our work. 

	\subsubsection{Surrounding vehicles}
	Surrounding vehicles' trajectories will be simulated based on a microscopic car-following model so as to reflect a realistic and dynamic traffic environment. The surrounding traffic information will get updated every $\tau_s=0.4$ seconds. At each updating step, four functions will be executed by surrounding vehicles in the following sequence: join/exit from the highway, merge into/split from a platoon, change lane, and adjust velocity based on the car-following model. These functions are elaborated in the following:
	
	\vspace{0.5pc}
	\noindent 1. \underline{Join/exit from the highway}:
	We assume that the probability that a vehicle enters the highway from an on-ramp at each updating epoch is $p_{\text{on}}$. The vehicle is assumed to be able to join the highway if it can maintain a minimum time gap of length $t_p$ from the vehicles both upstream and downstream of the ramp entry point in the right lane of the highway.
	We set the speed of this entering vehicle similar to the speed of its downstream vehicle. Moreover, we set the probability of the vehicle not being a platoon-enabled vehicle as $p_{\text{npe}}$.
	
	We assume that at each update step each vehicle may take an off-ramp with the probability $p_{\text{off}}$. A vehicle that is marked to leave the highway based on this probability may do so if it is traveling on the right lane of the highway, on the upstream of an off-ramp point, and the time gap between the vehicle and the off-ramp point is smaller than the update step, $\tau_s$. This exiting vehicle and its profile is directly taken off the current iteration.
	
	\vspace{0.5pc}
	\noindent 2. \underline{Merge into/split from a platoon}:
	It is assumed that a vehicle could hold only a single platoon membership status (either a member or not a member) throughout a road piece, i.e., the merging or splitting process can only commence in the transition point between two road pieces. (Note that that is the reason for assuming a $\beta$ value of 0.95 for the merge and split sub-actions in Eq. (\ref{eq:cost1}).) A vehicle can merge into a platoon when it is already a platoon leader (resulting in merging of two platoons), or a platoon-enabled free agent. Among all vehicles that qualify to merge into a platoon, the probability that a vehicle intends to merge is assumed to be $p_{\text{merge}}$. There are two cases regarding the profile of the vehicle in the immediate downstream of the merging vehicle. If it is a platoon member, then the new merging vehicle will have the same scheduled splitting position as other vehicles in the platoon. If it is a single vehicle, the scheduled splitting position $P_{\text{sch}}$, in units of number of road pieces, will be decided at this time using a normal distribution. (For more details, see section \ref{sec:PlatoonMembership}.)
	
	Every time when a platoon passes the transition point of two road pieces, the scheduled splitting position will decrease by 1 unit until this value reaches 0, at which point the platoon would split into free agents.
	
	\vspace{0.5pc}
	\noindent 3. \underline{Lane change}:
	\cite{rahman2013review} provides a comprehensive review of prior work on lane changing models. For simplicity, in this paper we adopt the random lane changing (RLC) model, in which vehicles may change lane once a minimum gap criterion is satisfied. We assume that in every update step at most a single vehicle can change lane. Furthermore, for safety considerations, we require a minimum time (no less than $t_{\textbf{lc}}=5$ seconds) between two successive lane changes by two successive vehicles (immediate follower/leader) traveling in the same lane. We allow only free agents, and not platoons, to change lane. The gap between the lane changing vehicle and surrounding vehicles (the leading vehicle in the same lane, and the leading and following vehicles in the target lane) should be large enough (larger than $d_{\textbf{cg}}$) to ensure a safe lane changing maneuver. Finally, the following vehicle in the target lane cannot be a follower in a platoon, indicating that the lane changing process cannot insert vehicles into a platoon.
	
	Not all vehicles that satisfy the conditions above intend to change lane. Among all qualified vehicles, the probability that a vehicle intends to change lane is $p_{\text{change}}$. The lane changing process is assumed to be completed within time $t_{\text{lcp}}$ seconds, after which the lateral position of the lane changing vehicle would not change, and its longitudinal speed has to reach the speed of the leading vehicle in the target lane.
	
	\vspace{0.5pc}
	\noindent 4. \underline{Adjusting velocity using a car-following model}:
	%It is assumed that the first vehicle in each lane will not adjust their velocity (i.e., they can have any speed smaller that the free-flow speed). Other 
	Each vehicle needs to continuously adjust its velocity to maintain a large enough safety gap from its leading vehicle. The adjustment of velocity depends on the platoon membership status of the vehicle. We use the Intelligent Driver Model (IDM) \cite{jin2016optimal} for adjusting velocity. The parameters used to calibrate the IDM are summarized in the Table \ref{Parameters summary}.
	
	\subsubsection{Subject vehicle}
	%We adjust the subject vehicle's controller to simulate how different controllers may influence fuel consumption by the subject vehicle and its surrounding vehicles. 

    \begin{figure*}[h]
    \centering\includegraphics[scale=0.4]{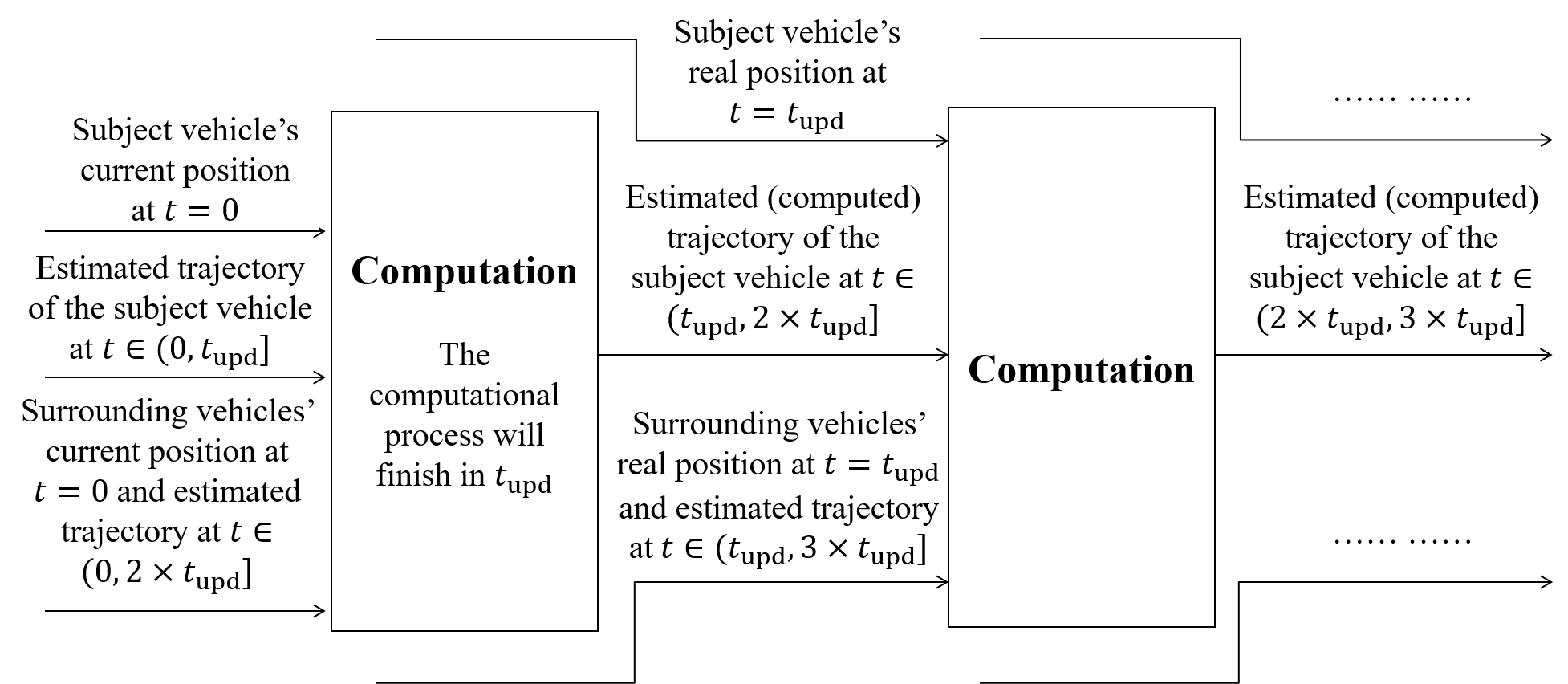}
    \caption{The computational process inputs and outputs: Each computational process for the optimal trajectory of the subject vehicle is finished within the time $t_{\text{upd}}$. The output of the last computational process serves as an input to the next computational process. Once a computational process is completed, the last trajectory of the subject vehicle is updated based on the new trajectory. }
    \label{fig:computation time delay}
    \end{figure*}

    %\paragraph{computation delay and integrated process}
    The subject vehicle updates its motion plan every $t_{\text{upd}}=0.4$ seconds. It is assumed that surrounding vehicles' motion information is available to the subject vehicle in real-time.
    Due to the long computational time of trajectory planning and control in a dynamic driving environment, it is problematic for the subject vehicle to obtain the latest traffic information and then plan its own trajectory for the immediate next period; that is, after the trajectory planning process is completed, the planned trajectory would be already outdated. Thus, we put forward a revised computational process in this paper, as shown in the Figure \ref{fig:computation time delay}. During this process, the subject vehicle perceives the environment, estimates other vehicles' motions for the next $2 t_{\text{upd}}$ period, and makes its own trajectory plan for the second following period, i.e., $[t+t_{\text{upd}}~,~t+2t_{\text{upd}}]$, where $t$ is the current time. This results in a trajectory that can still be effectively followed during this window. %It should be noted that even if the updating period of the trajectory for the subject vehicle is $t_{upd}=upd_{sub}=0.4$ seconds, the environmental information would be utilized for the next $2 \item t_{upd}$ period, which indicates that the computation time delay is in fact $2*t_{upd}=0.8$ seconds. The relationship between time delay and stability or collision avoidance is out of the scope of this paper.

 As discussed in \cite{Luo2016A}, the subject vehicle may get involved in a collision due to the surrounding vehicles' sudden speed fluctuations during the lane changing process. More specifically, the subject vehicle may not be able to take any action without violating the constraints of the optimal control model for the following reasons: ($i)$ the sudden speed change of the surrounding vehicles; ($ii$) the comfort-related maximum acceleration and jerk constraints in the optimal control model; and ($iii$) the conservative constraints regarding the safety time gap between the subject vehicle and any surrounding vehicles. In case of there being no feasible solution for the optimal control model, the Intelligent Driver car-following Model is utilized to provide a longitudinal motion reference for the subject vehicle.
% %As for the lateral motion during the lane changing process, the subject vehicle has two choices: go back to the original lane or keep the lateral speed as it is in the last updating period. Here, we assume that the subject vehicle keeps its lateral speed if the leading vehicle in the target lane is further away than the leading vehicle in original lane (i.e., the distance gap in the target lane is larger than the distance gap in the original lane); otherwise, we assume the subject vehicle go back to the original lane. In one word, the subject vehicle would choose the safest lane if no feasible solution was found in the optimal control model.
	
	\subsubsection{Platoon membership}\label{sec:PlatoonMembership}
	%We assume that all platoon-enabled vehicles can be platoon members. , the subject vehicle can also be platoon leader if the controller adopted supports that. 
	This section elaborates on platoon formations.
	When merging, we assume a free agent or a platoon can merge with its immediate downstream free agent or platoon. That is, merging can occur between two free agents, two platoons, or a free agent and a platoon. 
	We assume a finite number of possible scheduled splitting positions, $\ell_\text{sch}^1$, $\ell_\text{sch}^2$, $\cdots$, $\ell_\text{sch}^n$, in an ascending order of time. Given the mean $\mu_{\text{sch}}$ and the standard deviation $\sigma_{\text{sch}}$, we draw a random number $p_{\text{sch}}$ from the normal distribution $\mathcal{N}(\mu_{\text{sch}},\sigma_{\text{sch}})$ to schedule a splitting time, where $\ell_{\text{sch}}^{i-1}<p_\text{sch} \leq \ell_{\text{sch}}^{i}$ indicates selecting the scheduling time $P_\text{sch}=\ell_{\text{sch}}^{i}$. We set $P_{\text{sch}}=\ell_{\text{sch}}^n$ if $p_\text{sch} > \ell_{\text{sch}}^{n}$.
	%except that $P(psch_1)=P(Psch<=psch_1)$ and $P(psch_n)=1-P(Psch<=psch_{n-1})$.
	At the scheduled splitting position, platoon members will detach one by one, starting from the platoon tail, by increasing their gap with their immediate downstream vehicle.

	\begin{table*}[]
	\centering
\caption{Summary of parameters}
\label{Parameters summary}
\scalebox{0.9}{
\begin{tabular}{l|l|l}
\hline
Parameter                                           & Value                                                              & Definition                                                                                                   \\ \hline
$t_{\text{upd}}$                                           & 0.4 secs                                                           & updating period of trajectory of the subject vehicle                                                      \\ %\hline
$p_{\text{on}}$                                            & 0.6                                                                & \begin{tabular}[c]{@{}l@{}}the possibility of that a vehicle is interested\\ in joining the freeway from on-ramp \end{tabular}                       \\ %\hline
$p_{\text{off}}$                                           & 0.6                                                                & \begin{tabular}[c]{@{}l@{}}the possibility of that a vehicle is interested\\ in taking at off-ramp \end{tabular}                                  \\ %\hline
$p_{\text{npe}}$                                           & 0.5                                                                & \begin{tabular}[c]{@{}l@{}}the possibility of that the vehicle is a \\ non-platoon-enabled vehicle \end{tabular}                                    \\ %\hline
$p_{\text{merge}}$                                         & 0.6                                                                & the possibility of that a vehicle intends to merge                                                        \\ %\hline
$p_{\text{change}}$                                        & 0.1                                                                & the probability of that the vehicle intends to change lane                                                \\ %\hline
$t_{p}$                                             & 3.5 secs                                                           & \begin{tabular}[c]{@{}l@{}}time gap between two successive vehicles that\\ are not in a platoon \end{tabular}                                 \\ %\hline
$t_{g}$                                             & 0.55 secs                                                          & time gap between two successive vehicles in a platoon                                                     \\ %\hline
$t_{\text{lcp}}$                                             & 3.6 secs                                                          & surrounding vehicles finish lane changing within this time                                           \\ %\hline
$t_{\text{lc}}$                                            & 5 secs                                                             & \begin{tabular}[c]{@{}l@{}}the minimum time interval between two successive\\ lane changes by two successive vehicles in the same lane\end{tabular} \\ %\hline
$\tau_{s}$                                   & 0.4 secs                                                           & reaction time delay in the car-following model    \\ %\hline
$t_{N_{\text{act}}}$                                   & 10 secs                                                           & prediction horizon in optimal control model                                                            \\ %\hline
$v_{\text{m}}^{\text{le}}$                          & 20 m/s                                                             & \begin{tabular}[c]{@{}l@{}}the velocity in left lane when it reaches\\ the maximum flow\end{tabular}                                                \\ %\hline
$v_{\text{m}}^{\text{ri}}$                          & 14 m/s                                                             & \begin{tabular}[c]{@{}l@{}}the velocity in right lane when it reaches\\ the maximum flow\end{tabular}                                               \\ %\hline
$v_{\max}^{\text{le}}$                                      & 30 m/s                                                             & maximum velocity in left lane                                                                             \\ %\hline
$v_{max}^{\text{ri}}$                                      & 20 m/s                                                             & maximum velocity in right lane                                                                            \\ %\hline
$a_{\max}$                                    &  2 $\text{m}/{\text{s}}^\text{2}$                                                                   & maximum acceleration for subject vehicle                                                                  \\ %\hline
$j_{\max}$                                    &  3.5 $\text{m}/{\text{s}}^\text{3}$                                                                   & maximum jerk for subject vehicle                                                                          \\ %\hline
$d_{\text{cg}}$                                            & 50 m                                                               & \begin{tabular}[c]{@{}l@{}}critical gap decide whether it is feasible\\ to change lane\end{tabular}                                                  \\ %\hline
$l_{\text{car}}$                                         & 5 m                                                                & length of a vehicle                                                                                       \\ %\hline
$h_{\text{st}}$                                            & 5 m                                                                & vehicle would stop at headway of this value                                                               \\ %\hline
$a$                                                 & 2 $\text{m}/{\text{s}}^\text{2}$                                           & the maximum desired acceleration                                                                          \\ %\hline
$b$                                                 & 3 $\text{m}/{\text{s}}^\text{2}$                                           & the comfortable deceleration                                                                              \\ %\hline
$\gamma_{\text{AR}}$                                       & 0.3987                                                             & coefficient for air resistance force                                                                      \\ %\hline
$\gamma_{\text{RR}}$                                       & 281.547                                                            & coefficient for rolling resistance force                                                                  \\ %\hline
$\gamma_{\text{GR}}$                                       & 0                                                            & coefficient for grade resistance force                                                                  \\ %\hline
$\gamma_{\text{IR}}$                                       & 1750                                                               & coefficient for inertia resistance force                                                                  \\ %\hline
$\eta_f$                                            & \begin{tabular}[c]{@{}l@{}}5.98$\times\text{10}^{\text{-8}}$\\ dollars/Joule\end{tabular}                       & fuel cost for a unit energy consumed by the vehicle                                                       \\ %\hline
$P_{\text{sch}}$                                    & \{2,10,50\}                                                        & \begin{tabular}[c]{@{}l@{}}the scheduled splitting position can be\\ 2, 10 or 50 road pieces later\end{tabular}                                    \\ %\hline
$\mathcal{N}(\mu_{\text{sch}},\sigma_{\text{sch}})$ & 
\begin{tabular}[c]{@{}l@{}}$\mathcal{N}(2,5)$, left,\\ $\mathcal{N}(-1,5)$, right\end{tabular}
& \begin{tabular}[c]{@{}l@{}}the norm distribution of the scheduled splitting position\\ in two lanes, respectively\end{tabular}                                         \\ \hline
\end{tabular}
}
\end{table*}

	\section{Numerical Experiments}\label{sec:NUMERICAL EXPERIMENTS}
	%As mentioned before, a road piece has similar macroscopic properties. We assume the subject vehicle travels through a number of road pieces in the right lane of highway. The traveled path is composed of 20 road pieces, most of which are 600 meters, except that the first, fourth, twelfth and eighteenth road pieces are 400, 300, 200 and 300 meters respectively. We also assume that There are two on-ramps that are on the starting point of the first and eighteenth road pieces, respectively. There are three off-ramps that are on the starting point of the fourth, twelve, and the destination of the whole trip, respectively.

	In this section we conduct experiments in the simulation framework laid out in the previous section, where the trajectory of the subject vehicle is controlled by the proposed optimal control model. The simulation framework consists of a two-lane highway, where the subject vehicle is assumed to be initially traveling on the right lane. The traveled path is composed of 20 road pieces, with two on-ramps in the first and eighteenth road pieces, and three off-ramps on the fourth, twelve, and the destination of the trip. The travel path is 10.8 km in length, where the first, fourth, twelfth and eighteenth road pieces are 400, 300, 200 and 300 meters in length, respectively, and the rest of the road pieces are 600 meters in length. Recall that we consider a road piece to be homogeneous in macroscopic traffic conditions.

We quantify the implications of the optimal control model under different configurations of platooning (enabled or not) and lane changing (enabled or not), in different traffic environments. Specifically, we consider three traffic states, namely, free-flow traffic, onset-of-congestion traffic, and congested traffic.  
%In order to set a realistic initial traffic environment, traffic flow fundamental diagram is utilized for simulation in three different traffic situations (free-flow traffic, onset-of-congestion traffic and congested traffic). 
In order to provide a realistic simulation environment under each traffic state, we setup a warming-up process, during which we use a fundamental diagram of traffic flow to create simulation instances under each traffic state. According to \cite{qu2017stochastic}, many different models have been proposed to capture the relationship among the three fundamental parameters of traffic flow---traffic flow, speed, and traffic density. Here, we adopt Greenberg's model, which presents one of the earliest and most well-known speed-density models \cite{greenshields1935study,pipes1966car}. Let $v_{\text{m}}$ and $k_{\text{m}}$ be the corresponding velocity and density when the flow reaches its maximum value, which is $\frac{1}{t_{\text{p}}}$. We set $k_{\text{1}}=0.3\: k_{\text{m}}$, $k_{\text{2}}=0.8\: k_{\text{m}}$, and $k_{\text{3}}=2 \: k_{\text{m}}$ as the maximum density under the free-flow, onset-of-congestion, and congested traffic states, respectively. We then use Greenberg's speed-density relationship in Eq. (\ref{speed-density}) to compute the corresponding velocity of each of the three density cut-off points,
%For a given traffic state, one representative point (density, speed) is chosen to set microscopic variables (speed, headway).
\begin{equation}\label{speed-density}
v=v_{\text{m}}\:\ln\Big(\frac{k_{\text{j}}}{k}\Big)
\end{equation}
where $v$ denotes the space-mean-speed, $k$ denotes the traffic density, $v_{\text{m}}$ indicates the velocity when the flow reaches its maximum value, and $k_{\text{j}}$ indicates the jam density. Value of $k_{\text{j}}$ is determined by the parameters in the IDM model,
%\begin{equation}\label{v_f}
%v_{\text{f}}=v_{\text{max}}
%\end{equation}
\begin{equation}\label{k_j}
k_{\text{j}}=\frac{1}{l_{\text{car}}+h_{\text{st}}}%=\frac{1}{10}
\end{equation}
where $l_{\text{car}}$ is the average vehicle length, and $h_{\text{st}}$ is the minimum headway at which vehicles are at a complete stop. After generating vehicle positions using the ideal time gap, we perturb these positions using Gaussian noise to incorporate random deviations from an idealized model. During the warm-up process all surrounding vehicles run for 2 minutes following the IDM model.

	%We compare the fuel costs of the subject vehicle resulted from behaviors like car-following, optimal control and platoon joining in different traffic situations. For each situations, we simulated 25 times with randomly generated initial traffic environment, the box-plot of all simulations are shown in figure \ref{fig:result1}. 

	For each traffic state, we run seven simulation scenarios, each scenario using a different controller for the subject vehicle, as follows: (1) the subject vehicle travels according to the intelligent driver car-following model \cite{jin2016optimal} and cannot join a platoon (CF), (2) the subject vehicle travels according to the optimal control model and cannot join a platoon (OC), (3) the subject vehicle travels according to the optimal control model and can join a platoon; however, the platoon can dissolve at any point in time after formation (OC\_M0), (4) the subject vehicle travels according to the optimal control model and can join a platoon; however, a platoon has to travel for at least 6 km before it can dissolve (OC\_M6), (5-7) the subject vehicle can also change lane, in addition to the description in (2-4), respectively. For each traffic state, we run 25 random instances of each simulation scenario and report the trip cost, which is a linear combination of the fuel and time costs. 
	
	\subsection{Efficiency Results for the Subject Vehicle}\label{sec:subject}

\begin{figure*}[t]
    \centering\includegraphics[scale=0.4]{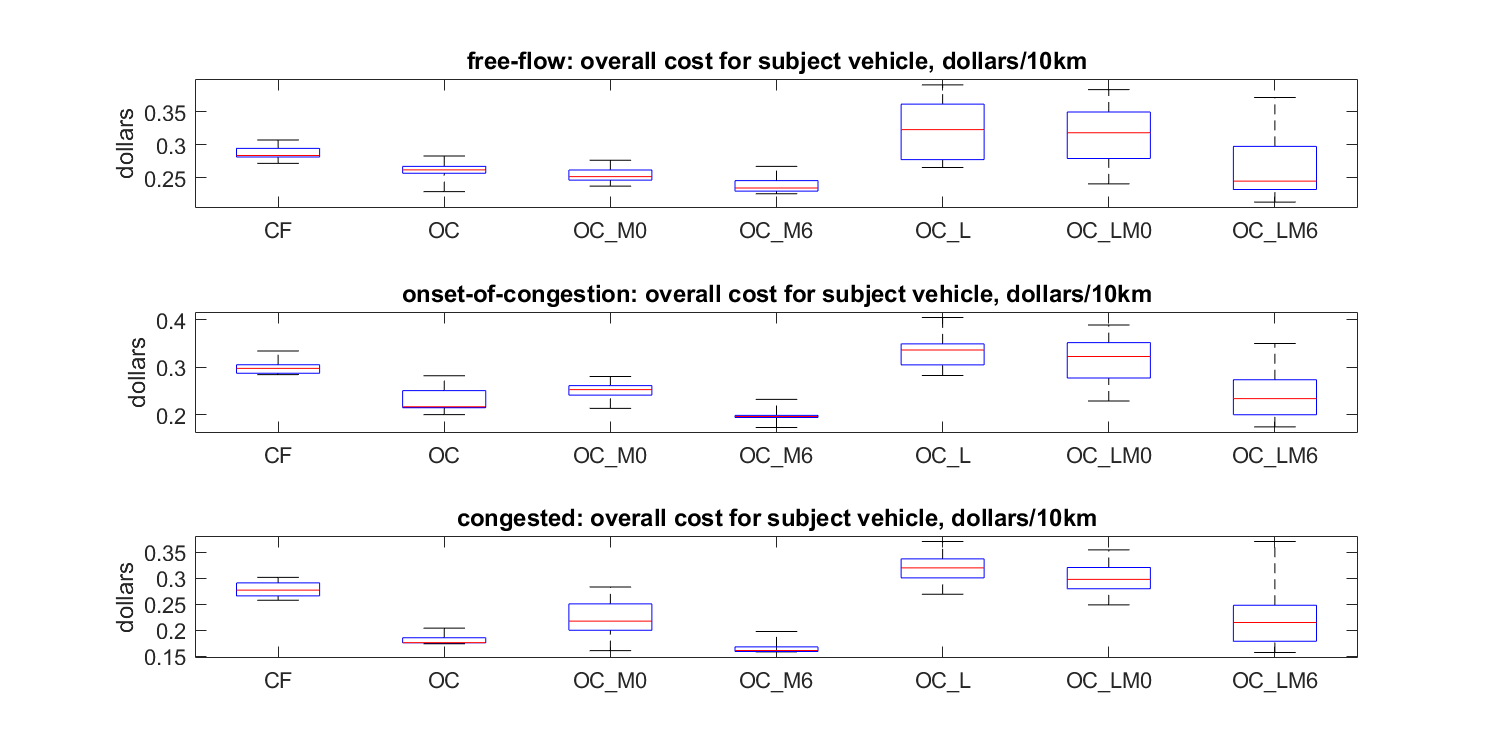}
    \caption{The top, middle, and bottom figures represent the free-flow, onset-of-congestion, and congested traffic states, respectively. The vertical axes in these figures show the overall cost in dollars for 10 km long trips. Along the horizontal axes, the overall costs of the subject vehicle under different controllers are compared. Value of time is set to 0 dollars per hour in all simulations shown in this figure. Here `CF' and `OC' denote the car-following and the optimal control models, respectively, where platooning is disabled. `OC\_Mx' indicates that the subject vehicle is enabled to merge into a platoon using the optimal control model, and that it needs to keep staying in the platoon for at least x km after the platoon is formed. `OC\_L', `OC\_LM0' and `OC\_LM6' indicate that lane changing is enabled on `OC', `OC\_M0' and `OC\_M6', respectively.}
    \label{fig:sub0}
    \end{figure*}

	In this section we report the overall cost of the subject vehicle under the seven introduced controllers, the three traffic states, and two different values of time. Figure \ref{fig:sub0} displays the results for the value of time $\eta_t=0$ dollars per hour, effectively comparing the fuel efficiency benefits of the seven controllers. 
	%To facilitate the comparison of scenarios, the results of the first scenario (i.e., CF) is repeated next to the three other scenarios. 
	The values of the overall fuel consumption by the subject vehicle under all scenario pairs are compared using a two-tailed Student's t-tests at the 5\% significance level to identify fuel savings that are statically significant.

    \begin{figure*}[t]
    \centering\includegraphics[scale=0.4]{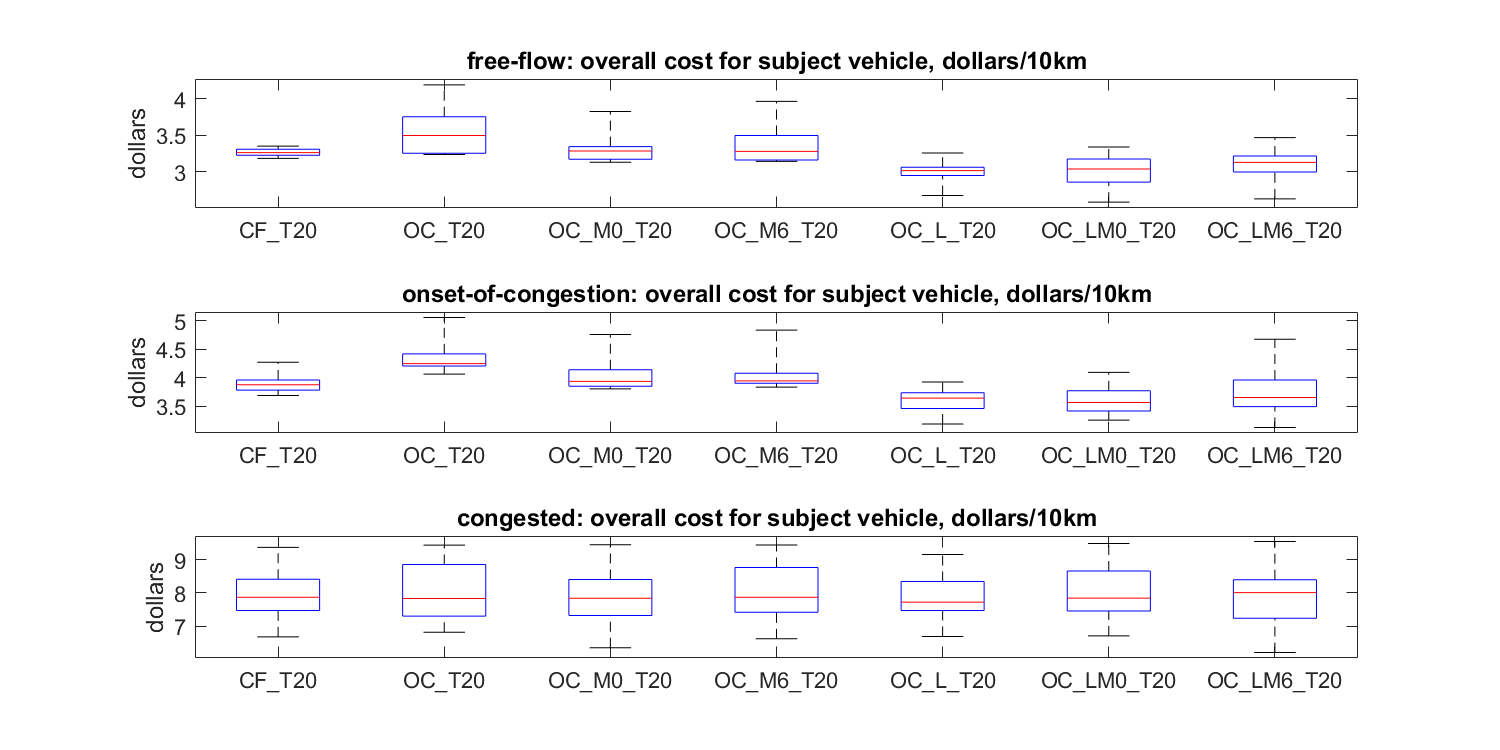}
    \caption{The suffix '\_T20' denotes that the value of time is set to 20 dollars per hour in all simulations shown in this figure. Other settings are the same with Fig. \ref{fig:sub0}.}
    \label{fig:sub20}
    \end{figure*}
	
	The top plot in Figure \ref{fig:sub0} presents the results for the free-flow traffic state. These results suggest that, without lane changing, the optimal control model, both with and without the ability to form a platoon, (that is, OC, OC\_M0, and OC\_M6) can result in statistically significant reductions in fuel cost (at the 5\% significance level), compared to the car-following model (CF). With lane changing, OC\_L and OC\_LM0 result in even higher fuel costs compared with CF. This is because the lane changing process itself may add to the fuel cost--a cost that might be underestimated by the short-sighted optimal control model. In general, if the subject vehicle is platoon-enabled and forced to keep its platoon membership for at least 6km (i.e., the OC\_M6 and OC\_LM6 scenarios), the fuel savings are more significant compared to OC alone. However, with lane changing, scenario OC\_LM0, where the platoon can dissolve at any point in time after its formation, does not produce statistically significant fuel savings compared to OC\_L. These results indicate that a stable, long-term platoon membership can have a positive effect on fuel efficiency.
	
	%The middle figure in Figure \ref{fig:result1} demonstrates the results for the  onset-of-congestion traffic state. Results indicate that similar to the free-flow case, optimal control offers statistically significant fuel savings compared to car-following. However, there is no statistically significant difference between the OC,  OC\_merge\_0km, and OC\_merge\_6km scenarios, suggesting that in the onset-of-congestion traffic state platooning does not offer fuel efficiency benefits. These results are not surprising since the onset-of-congestion traffic state could stochastically swing between free-flow and congested states during a short period of time, wiping out any potential benefits that platooning can offer beyond optimal control.
	
	The middle plot in Figure \ref{fig:sub0} demonstrates the results for the onset-of-congestion traffic state. Results indicate that similar to the free-flow case, without lane changing, optimal control offers statistically significant fuel savings compared to car-following for all control-based scenarios (with and without platooning). With lane changing, OC\_L results in higher fuel cost compared with CF, and OC\_LM0 has no significant difference with CF. However, comparison of OC, OC\_M0, and OC\_M6 scenarios in the onset-of-congestion traffic state shows that  OC\_M0 results in the least fuel saving, OC holds the second place, while OC\_M6 achieves the most energy saving. These results are not surprising since the frequent splitting of the subject vehicle from platoons in the onset-of-congestion state leads to higher energy consumption in the OC\_M0 scenario, and the energy savings from a short-lived platoon cannot make up for this loss.%the onset-of-congestion traffic state could stochastically swing between free-flow and congested states during a short period of time, wiping out any potential benefits that platooning can offer beyond optimal control.
	
	Finally, the bottom figure in Figure \ref{fig:sub0} displays the results for the congested traffic state. Results indicate that similar to the two previous traffic states, without lane changing, optimal control offers lower fuel cost compared to car-following. The OC\_M0 does not offer statistically significant improvements over the OC scenario for the same reason stated above; however, OC\_M6 can still offer statistically significant fuel savings over both OC and OC\_M0 controllers.
	
	In general, Figure \ref{fig:sub0} shows that regardless of traffic state, the OC model can outperform the CF model in terms of energy efficiency. Enabling platooning can increase these benefits even further if the model does not allow the platoon to dissolve at any point and enforces platoon members to travel together for a period of time. Lane changing could reduce the fuel efficiency benefits of the optimal control model to the point of matching fuel efficiency levels of traditional CF models; however, when platoon-keeping is enforced, the negative fuel efficiency implications of lane changing can be to negated to a great extent.
	
	%In summary, results indicate that OC\_M6 consistently produces more fuel-efficient trajectories for the subject vehicle compared to the OC\_M0 and OC, under all traffic states. A stable platoon membership can result in the most energy saving. While lane changing is enabled, OC\_LM6 also performs better than OC\_LM0 and OC\_L. However, OC\_L has a higher fuel cost compared with CF and OC. This is because vehicles in left lane usually have higher fuel costs, compared with vehicles in right lane. 
	%This can be attributed to the `shortsightedness' nature of the pure optimal control model, which only includes local information and optimizes over a prediction horizon when making platoon-merging and keeping decisions. 
	
	In Figure \ref{fig:sub20}, we set the value of time to be \$20 per hour and conduct simulations similar to those in Figure \ref{fig:sub0}. This figure shows that minimizing a generalized cost that takes into account the driver's value of time in addition to fuel cost turns lane changing into a more desirable feature of the optimal control model. 
	
	Under VoT of 20, in the congested traffic state there is no significant difference among all seven controllers. In the free-flow traffic state, we observe no statistically significant difference among OC\_L\_T20, OC\_LM0\_T20 and OC\_LM6\_T20, indicating that when lane changing is enabled platooning does not induce a significant change in the generalized cost. This is mainly due to the lower fuel cost compared to the high VoT. In onset-of-congestion traffic state, OC\_LM6\_T20 results in a slightly higher overall cost compared with OC\_L\_T20. It is due to the fact that when enforcing a platoon to hold for 6km, its members cannot change lane, resulting in a larger time cost. In free-flow and onset-of-congestion traffic state, different from Figure \ref{fig:sub0}, here the overall cost is reduced with lane changing. In both the free-flow and onset-of-congestion traffic state, OC\_M0\_T20 can result in significant overall cost savings compared with OC\_T20, while OC\_M6\_T20 has no significant difference compared with OC\_M0\_T20.
	
	By quantifying the effects of lane changing and platooning on the fuel and time costs, Figures \ref{fig:sub0} and \ref{fig:sub20} allow us to infer policies on the circumstances under which engaging in lane changing and/or platoon merging can reduce a vehicle's generalized trip cost. 
	In general, platooning reduces fuel cost and lane changing reduces the time cost of a trip. As such, the overall generalized cost becomes dependent on the relative values of VoT and fuel cost--if the value of time is small compared to the fuel cost, the contribution of platooning to the generalized cost overweighs that of the time cost, indicating a cost-minimizing policy of merging into platoons, committing to them for long periods, and avoiding lane changes. On the other hand, if VoT is large relative to the fuel cost, the time cost of the generalized cost overweighs the fuel cost, resulting in the cost-minimizing policy of not blindly committing to a platoon for a long period, while taking advantage of lane changing to reduce travel time when possible.

	%--------------------------------------------------
    \subsection{Efficiency Results for the Surrounding Vehicles}
	%--------------------------------------------------
	In this section, we analyze the simulation results to investigate whether the different controllers used by the subject vehicle have a significant impact on the overall cost of its upstream traffic. We use the average cost of $N_{\text{sur}}=30$ upstream vehicles of the subject vehicle in both lanes as an approximation of the cost of a surrounding vehicle. We assume that surrounding vehicles have the same value of time as the subject vehicle.

	\begin{figure*}[h]
    \centering\includegraphics[scale=0.4]{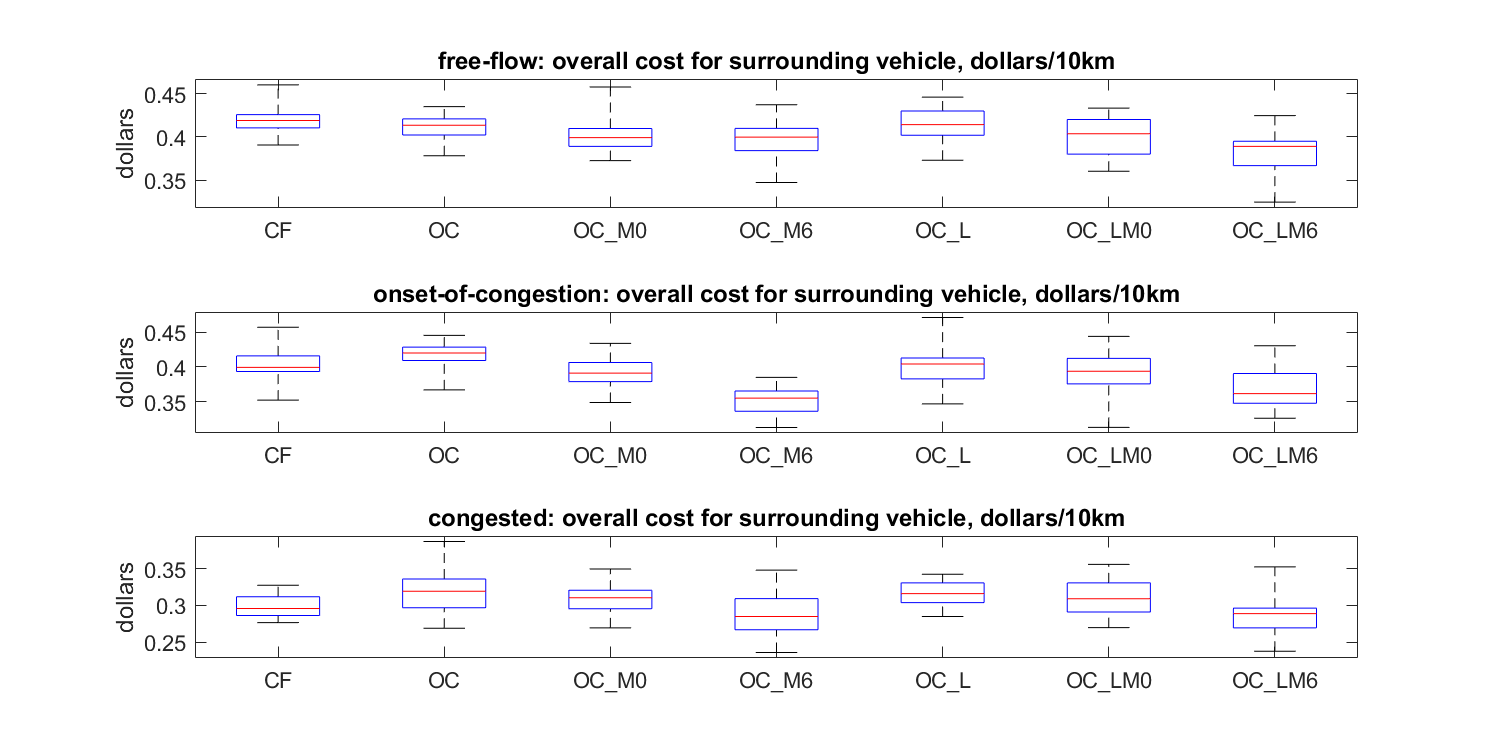}
    \caption{The vertical axis shows the overall cost per distance, the unit is dollars per 10 kilometers. Along the horizon axis, these cases that the subject vehicle is with different controllers are compared. Here 'CF', 'OC', 'OC\_Mx' 'OC\_L' and 'OC\_LMx' have the same meaning as in FIGURE \ref{fig:sub0}. 30 vehicles (15 vehicles in left lane, 15 vehicles in right lane) in the upstream of the subject vehicle are considered respectively, their overall costs are summed and averaged. Value of time is set to 0 dollars per hour in all simulations shown in this figure. We also considered three traffic situations for surrounding vehicles' overall costs.}
    \label{fig:sur0}
    \end{figure*}
    
    \begin{figure*}[h]
    \centering\includegraphics[scale=0.4]{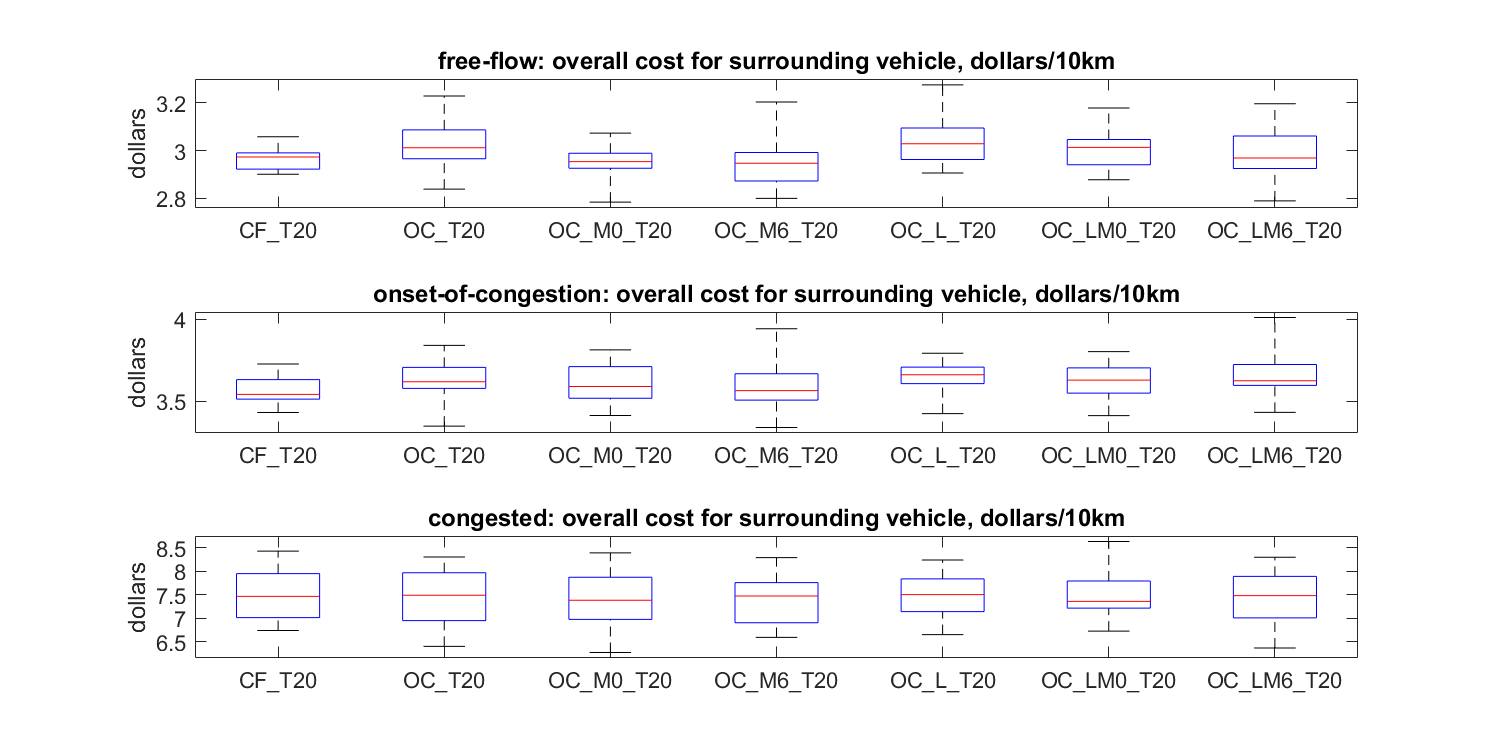}
    \caption{The suffix '\_T20' denotes that the value of time is set to 20 dollars per hour in all simulations shown in this figure. Other settings are the same with Fig. \ref{fig:sur0}.}
    \label{fig:sur20}
    \end{figure*}

    Figure \ref{fig:sur0} displays the average cost of $N_{\text{sur}}=30$ upstream vehicles to the subject vehicle under the three traffic states and the seven controllers, with value of time set to 0, thereby effectively measuring the impact of the controllers on fuel efficiency. This figure suggests that changing the subject vehicle controller from the car-following model to the optimal control model may have different implications in fuel consumption of the upstream vehicles depending on the traffic state. More specifically, replacing CF with OC results in significant fuel savings for the surrounding vehicles in the free-flow traffic, does not introduce a significant change in the onset-of-congestion traffic state, and induces a significant rise in fuel increasing under the congested traffic state. However, OC\_M6 performs better than CF, OC and OC\_M0 in all three traffic states, indicating that a connected vehicle can create fuel efficiency for its upstream traffic if it joins a platoon and commits to it. Similarly, when lane changing is enabled, OC\_LM6 outperforms OC\_L and OC\_LM0. Among all controllers, OC\_LM6 results in the most overall fuel savings for the surrounding vehicles.
    Finally, the subject vehicle's lane changing decisions do not create a significant difference in the surrounding vehicles' fuel consumption.

	In Figure \ref{fig:sur20}, we set the value of time to \$20 per hour. There is no statistically significant difference among controllers in the onset-of-congestion and congested traffic states. In the free-flow traffic state, OC\_T20 and OC\_L\_T20 result in larger cost for surrounding vehicles.

	\subsection{Impact of Platooning}
	
\begin{figure*}[h]
    \centering\includegraphics[scale=0.47]{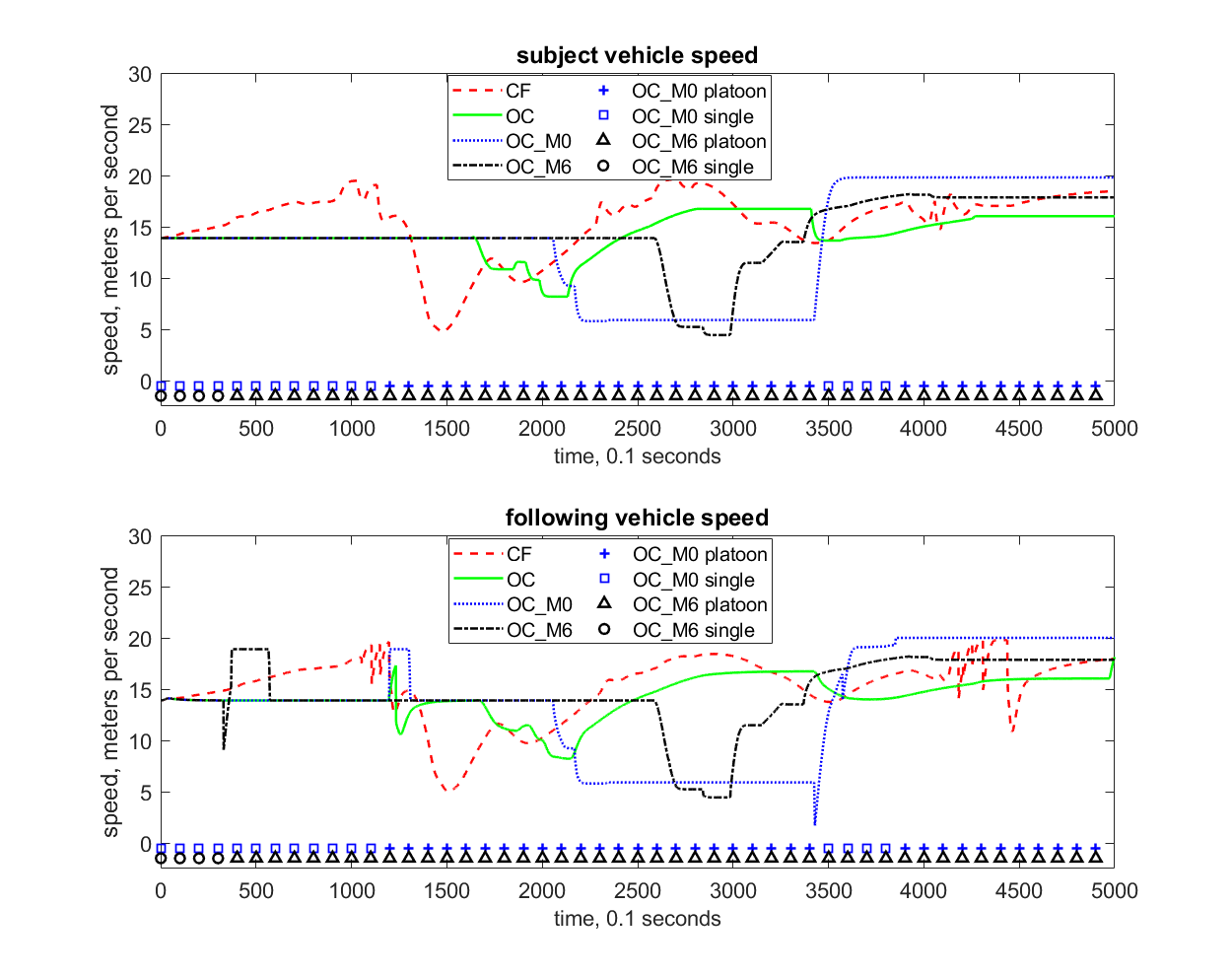}
    \caption{The vertical axis shows velocity, with the unit of meters per second. The horizontal axis is time, with a unit of 0.1 seconds. The top plot compares the speed curves of the subject vehicle under different controllers, and the bottom plot shows the corresponding speed curves of the immediate upstream vehicle to the subject vehicle. Here 'CF', 'OC' and 'OC\_Mx' have the same meaning as in Figure \ref{fig:sub0}. The platoon membership status of the subject vehicle is presented along the horizontal axis for the OC\_M0 and OC\_M6 controllers.}
    \label{last figure}
    \end{figure*}

	Figure \ref{last figure} allows us to pinpoint the source of fuel efficiency induced by the OC model. This figure shows the velocity curves of the subject vehicle and its immediate upstream vehicle in the onset-of-congestion traffic state in an example trip with VoT of 0. The points at the bottom of the plots in this figure mark the platoon membership status of the subject vehicle under the OC\_M6 and OC\_M0 controllers at each time epoch. In Figure \ref{last figure}, only the first 500 seconds of the trip are presented, and the fuel costs for this 500-second-long section of the trip as well as the entire trip are computed and shown in Table \ref{last fuel cost}. This figure shows that, compared to CF, OC provides smoother velocity trajectories, thereby resulting in fuel savings for both the subject vehicle and its immediate upstream vehicle. This figure also demonstrates that the OC\_M6 controller provides the smoothest trajectories, and therefore can provide the highest fuel-saving benefits.
	
	% Please add the following required packages to your document preamble:
% \usepackage{multirow}
\begin{table}[]
\centering
\caption{Fuel cost for subject vehicle and its immediate upstream vehicle in an example trip under the onset-of-congestion traffic state}
\label{last fuel cost}
\begin{tabular}{c|c|c|c|c}
\hline
\multirow{2}{*}{\begin{tabular}[c]{@{}c@{}}fuel cost,\\ dollars per 10 km\end{tabular}} & \multicolumn{2}{c|}{First 500 seconds}       & \multicolumn{2}{c}{The entire trip}          \\ \cline{2-5} 
                                                                                        & subject vehicle & following vehicle & subject vehicle & following vehicle \\ \hline
CF                                                                                      & 0.3096          & 0.3519                     & 0.3045          & 0.3518                     \\ %\hline
OC                                                                                      & 0.2420          & 0.2592                     & 0.2422          & 0.2753                     \\ %\hline
OC\_M0                                                                          & 0.2439          & 0.2614                     & 0.2424          & 0.2556                     \\ %\hline
OC\_M6                                                                          & 0.2238          & 0.2431                     & 0.2166          & 0.2295                     \\ \hline
\end{tabular}
\end{table}

	\subsection{Lane Changing and its Impact}
	
\begin{figure*}[h]
    \centering\includegraphics[scale=0.4]{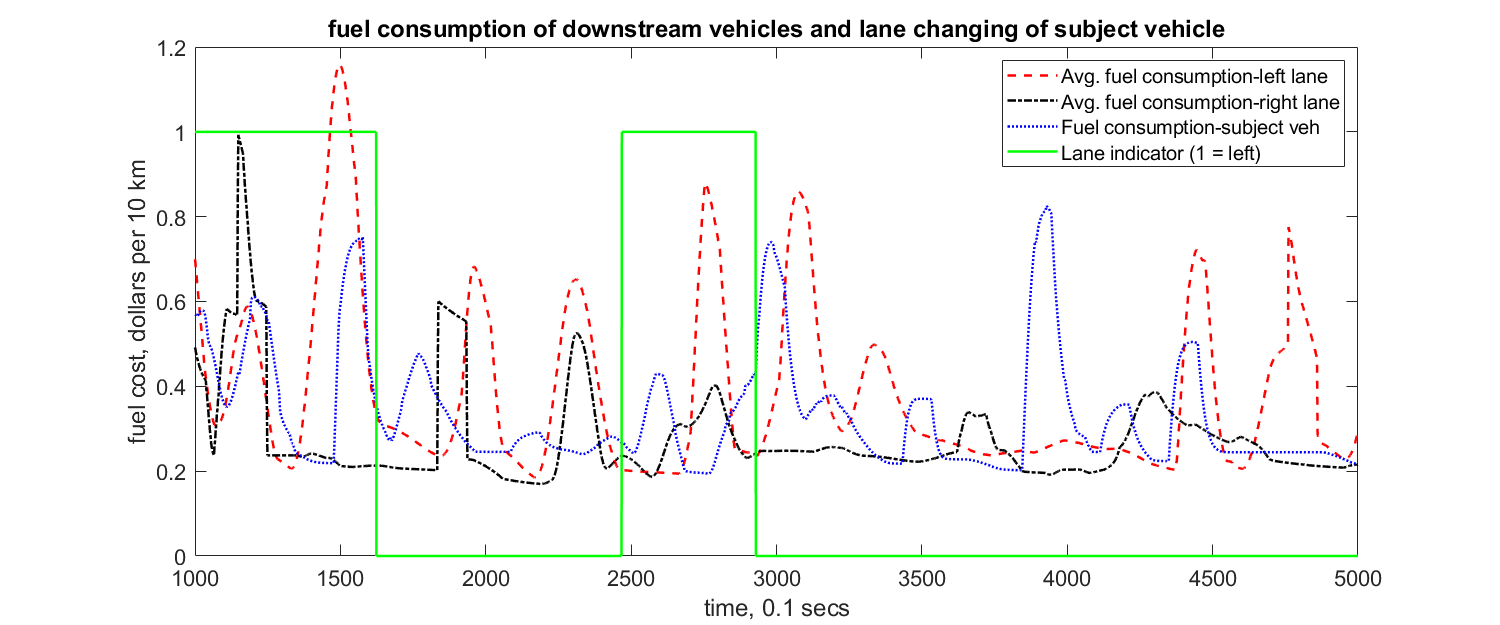}
    \caption{The vertical axis shows the fuel consumption, with a unit of dollars per 10 km. The horizontal axis is time, with the unit of 0.1 seconds. Average fuel cost of vehicles in downstream of the subject vehicles in both lanes (15 vehicles in left lane, 15 vehicles in right lane), and the fuel cost of subject vehicle and its lane position are shown.}
    \label{18down}
    \end{figure*}

	Figure \ref{18down} allows us to demonstrate how the subject vehicle makes lane changing decisions. This figure shows the fuel consumption curves of the subject vehicle and those of its downstream vehicles (averaged over 15 vehicles) on both the right and left lanes for an example trip in the onset-of-congestion traffic state. The controller of the subject vehicle is set to OC\_L. The green line indicates the lane in which the subject vehicle travels at each point in time, where the value 1 indicates the left lane. At about 1600 time epochs, the subject vehicle changes from the left lane to the right lane. This lane change can be attributed to the lower fuel consumption of downstream traffic in the right lane at about 1400 to 1600 time epochs. At about 2450 time epochs, the subject vehicle changes from the right lane to the left lane due to the lower fuel consumption of downstream traffic in the left lane at about 2450 to 2600 time epochs. The subject vehicle again switches from the left lane to the right lane at about 2900 time units due to the lower fuel consumption in the right lane at about 2750 to 2900 time units. As this figure shows, changing lane in response to reductions in fuel consumption in the other lane may bring upon short-term fuel savings, but the frequency of these lane changes may increase the total fuel cost, as was demonstrated and discussed in section \ref{sec:subject}.

\begin{figure*}[h]
    \centering\includegraphics[scale=0.4]{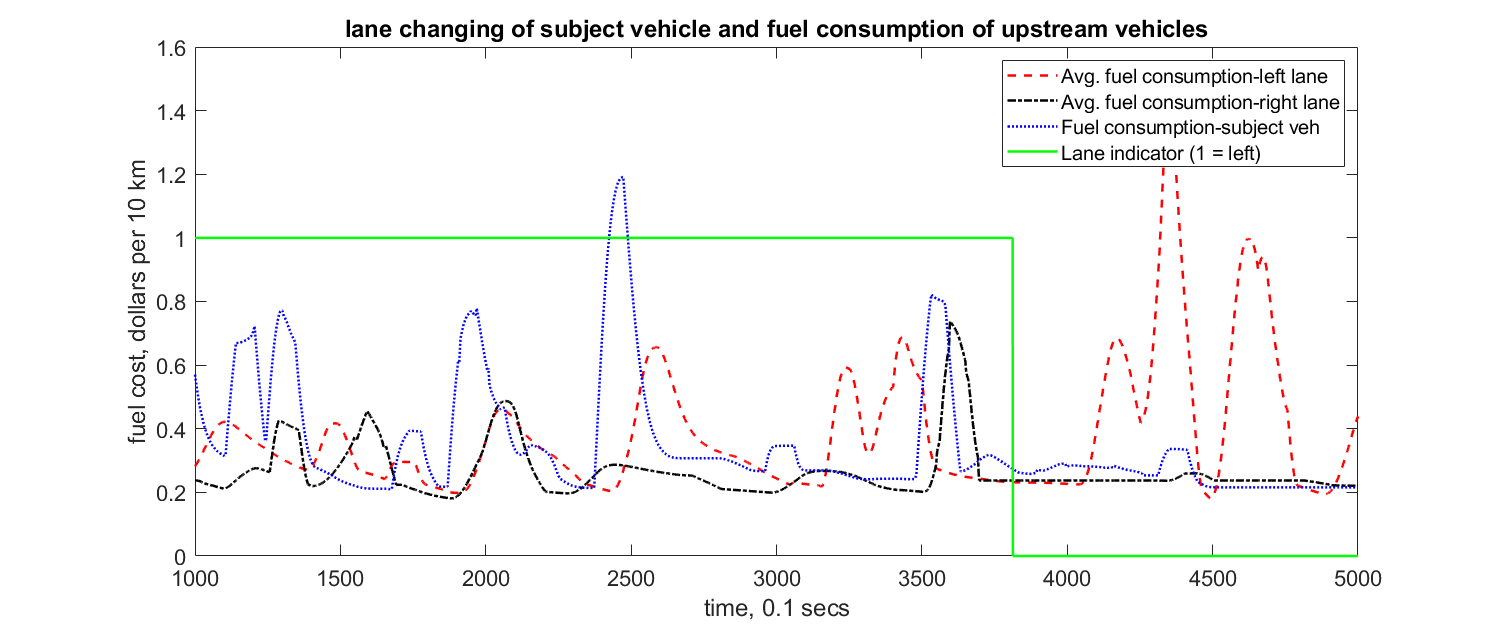}
    \caption{The vertical axis shows fuel consumption, with unit of dollars per 10 km. The horizontal axis is time, with unit of 0.1 seconds. Average fuel cost of vehicles in upstream of the subject vehicle in both lanes (15 vehicles in left lane, 15 vehicles in right lane), the fuel cost of subject vehicle and its lane position are shown.}
    \label{2up}
    \end{figure*}

    Figure \ref{2up} shows how the subject vehicle's lane changing decisions can influence the fuel consumption of upstream traffic in both lanes. This figure shows the fuel consumption curves of the subject vehicle and its upstream vehicles (in both lanes) in an example trip under the onset-of-congestion traffic state. The controller of the subject vehicle and the lane indicator are the same as in Figure \ref{18down}. At about 3800 time units, the subject vehicle changes from the left lane to the right lane. Figure \ref{2up} shows that the subject vehicle switching to the right lane does not negatively affect the fuel consumption in that lane, explaining the general trends in Figure \ref{fig:sur0}. 
    
    %, and we can see this results in that the right lane upstream traffic has less fuel consumption since then, and the left lane upstream traffic has more fuel consumption since then. This figure matches the intuition, the subject vehicle can benefit the following vehicles in the same lane.

\section* {Acknowledgement}
\noindent The work described in this paper was supported by research grants from the National Science Foundation (CPS-1837245, CPS-1839511).    
	
	\section{Conclusion}
	
	In this paper we proposed an optimal control model for trajectory planning of a CAV in a mixed traffic environment. The optimal controller was developed to plan the trajectory of the subject vehicle, including platoon formation and lane changing decisions, while explicitly accounting for computation delay. The objective of the optimal control model is to minimize a generalized cost that is a linear combination of fuel and time costs. We developed a simulation framework to quantify the effectiveness of the optimal control model in providing first-hand energy savings for the subject vehicle as well as second-hand energy savings for the vehicles traveling upstream of the subject vehicle. Our experiments suggest that, generally speaking, if the value of time of the driver is 0 dollars per hour--reducing the generalized cost to the fuel cost--, the optimal controller performs better than the IDM car-following model. Results suggest that making platooning decisions based on local information does not necessarily lead to fuel savings; however, if a minimum platoon-keeping distance is imposed by the model, platooning can offer significant fuel-efficiency benefits, specially in the onset-of-congestion and congested traffic states. Our experiment also indicate that, under the controller with minimum platoon-keeping distance requirement, the non-connected vehicles upstream of the subject vehicle may also experience second-hand statistically-significant fuel savings. When value of time is set to 20 dollars per hour, lane changing may introduce time savings significant enough to more than compensate the increased fuel consumption during the lane change maneuver, and in fact reduce the overall cost of the trip. As such, our experiments indicate the importance of the relative values of fuel cost and value of time in drivers' decision-making process--with higher value of time, lane changing becomes more attractive, leading to the generalized cost preferring a shorter trip to a more fuel-efficient one. Similarly, with a smaller value of time one might prefer to engage in platoon formation to reduce his/her fuel cost. This interesting relationship can open doors for introducing mechanisms between agent where those with lower value of time might grant lane access to those with higher value of time for a monetary compensation, thereby increasing utilities of all parties.%Furthermore, we investigate the decision making of lane changing and its impact. We notice it may change lane multiple times in some scenarios, because the optimization is conducted locally. Finally, simulation results under different value of time match our intuition, when value of time is small, platooning can play a big role in reducing overall cost and lane changing is not encouraged. When value of time is big, lane changing can save travel time, while platooning means staying with surrounding vehicles in one lane, which cannot reduce overall cost. The balance point that we found in our experiment setting is $VoT=5$ dollars per hour, platooning and its maintaining distance cannot make a significant difference.

\bibliographystyle{chicago}
\bibliography{refs}

%\section*{Appendix I} 

\end{document}